\documentclass[sigconf]{acmart}

\usepackage{multirow} 
\usepackage{algorithm}  
\usepackage{algpseudocode}  
\usepackage{amsmath}  
\usepackage{subfigure}
\usepackage{graphicx}

\AtBeginDocument{%
  }

\copyrightyear{2025}
\acmYear{2025}
\setcopyright{cc}
\setcctype{by}
\acmConference[WWW '25]{Proceedings of the ACM Web Conference 2025}{April 28-May 2, 2025}{Sydney, NSW, Australia}
\acmBooktitle{Proceedings of the ACM Web Conference 2025 (WWW '25), April 28-May 2, 2025, Sydney, NSW, Australia}
\acmDOI{10.1145/3696410.3714765}
\acmISBN{979-8-4007-1274-6/25/04}

\begin{document}

\title{Division-of-Thoughts: Harnessing Hybrid Language Model Synergy for Efficient On-Device Agents}

\author{Chenyang Shao}
\affiliation{%
 \institution{Department of Electronic Engineering\\ BNRist, Tsinghua University}
 \city{Beijing}
 \country{China}}
 \email{shaocy24@mails.tsinghua.edu.cn}

\author{Xinyuan Hu}
\affiliation{%
 \institution{Department of Quantitative Theory \& Methods\\Emory University}
 \city{GA}
 \country{USA}}
 \email{nate.hu@emory.edu}

\author{Yutang Lin}
\affiliation{%
 \institution{Department of Electronic Engineering \\Tsinghua University}
 \city{Beijing}
 \country{China}}
 \email{yt-lin21@mails.tsinghua.edu.cn}

\author{Fengli Xu$^{*}$}
\affiliation{%
 \institution{Department of Electronic Engineering\\ BNRist, Tsinghua University\authornote{Corresponding author.}}
 \city{Beijing}
 \country{China}}
 \email{fenglixu@tsinghua.edu.cn}

\begin{abstract}
The rapid expansion of web content has made on-device AI assistants indispensable for helping users manage the increasing complexity of online tasks. The emergent reasoning ability in large language models offer a promising path for next-generation on-device AI agents. However, deploying full-scale Large Language Models (LLMs) on resource-limited local devices is challenging. In this paper, we propose \underline{D}ivision-\underline{o}f-\underline{T}houghts (\textbf{DoT}), a collaborative reasoning framework leveraging the synergy between locally deployed Smaller-scale Language Models (SLMs) and cloud-based LLMs.
DoT leverages a \textit{Task Decomposer} to elicit the inherent planning abilities in language models to decompose user queries into smaller sub-tasks, which allows hybrid language models to fully exploit their respective strengths. Besides, DoT employs a \textit{Task Scheduler} to analyze the pair-wise dependency of sub-tasks and create a dependency graph, facilitating parallel reasoning of sub-tasks and the identification of key steps. To allocate the appropriate model based on the difficulty of sub-tasks, DoT leverages a \textit{Plug-and-Play Adapter}, which is an additional task head attached to the SLM that does not alter the SLM's parameters. To boost adapter's task allocation capability, 
we propose a self-reinforced training method that relies solely on task execution feedback.
Extensive experiments on various benchmarks demonstrate that our DoT significantly reduces LLM costs while maintaining competitive reasoning accuracy. Specifically, DoT reduces the average reasoning time and API costs by 66.12\% and 83.57\%, while achieving comparable reasoning accuracy with the best baseline methods. \footnote{Code available at: https://github.com/tsinghua-fib-lab/DoT}
\end{abstract}

\ccsdesc[500]{Computing methodologies~Natural language generation}
\ccsdesc[300]{Computer systems organization~Client-server architectures}

\keywords{Large Language Model, LLM Reasoning, AI Agents, Edge-Cloud Collaboration}


\maketitle

\section{Introduction}

As web content continues to grow exponentially, on-device AI assistants have become essential tools for helping users navigate the increasingly complex online landscape. This trend has led to the widespread adoption of personal assistants such as Google Assistant, Apple Siri, Amazon Alexa, Alibaba Tmall Genie, and Xiaomi Xiao AI~\cite{kepuska2018next,mari2019voice}, which have demonstrated their effectiveness in helping users digest enormous web content for tasks like web browsing~\cite{lai2024autowebglm, chen2024large, zhou2024synergizing}, content searches~\cite{sharma2024generative}, online shopping~\cite{nie2024hybrid}, and travel planning~\cite{shao2024beyond, chiu2009towards}. These AI-powered agents enable web applications to harness the rapid advancements in AI technology, delivering a more personalized and convenient user experience. Amid recent AI breakthroughs in Large Language Models (LLMs), the emergent capabilities of commonsense reasoning~\cite{wei2022chain} and in-context learning~\cite{dong2022survey} are widely regarded as a key component for the next generation of on-device agents~\cite{gunter2024apple}. Therefore, revolutionizing AI personal assistants with LLM agents has become an important research problem and a critical focus for applications~\cite{li2024personal, li2024limp}.

However, deploying LLM agents on local devices presents significant challenges, as it is impractical to run LLMs with trillions of parameters on resource-constrained devices such as smartphones and personal computers~\cite{xu2024survey}. Conversely, relying solely on cloud-based commercial LLMs raises concerns over privacy risks, unreliable connections, and high monetary costs~\cite{li2024personal}. Recent research has focused on training smaller-scale language models and developing model compression techniques~\cite{xu2024device,lu2024small,lin2024awq}, with the goal of creating sub-10B parameter models that can be practically deployed on local devices, such as Llama 3 series~\cite{dubey2024llama} and Phi 3 series~\cite{abdin2024phi}. However, this approach introduces additional computational costs for training or compressing these models and inevitably results in performance degradation compared to full-size LLMs. Preliminary efforts have also explored the potential of edge-cloud collaboration, where tasks exceeding the capabilities of locally deployed \underline{S}maller-scale \underline{L}anguage \underline{M}odels (SLMs) are rerouted to more powerful cloud-based LLMs~\cite{chen2023frugalgpt,gupta2024language}. Despite this, the significant performance gap between SLMs and LLMs often leads to a suboptimal trade-off between reasoning capabilities and cost.

To address this problem, our work draws inspiration from the fundamental economic concept of ``division of labour'', which posits that breaking down complex tasks into finer components often leads to more efficient solutions by allowing collaborative partners to fully exploit their respective strengths. Building on this idea, we propose a novel \underline{D}ivision-\underline{o}f-\underline{T}houghts (\textbf{DoT}) framework to fully harness the synergy between locally deployed SLMs and cloud-based LLMs through sophisticated task decomposition and optimized sub-tasks allocation. Specifically, DoT leverages a \emph{Task Decomposer}, a meta-prompt that combines ``chain-of-thought''-like prompting~\cite{wei2022chain} with carefully curated task decomposition examples. This approach taps into the inherent planning abilities of language models~\cite{song2023llm}, enabling them to decompose user queries into smaller sub-tasks. Our core insight is that even complex user queries often contain a significant portion of simple sub-tasks that can be adequately handled by SLMs. Therefore, decomposing user queries can lead to optimized collaboration on sub-task level, i.e., unleashing the potential of “Division-of-Thoughts”. Moreover, DoT employs a \emph{Task Scheduler} that analyzes the pair-wise dependency between sub-tasks and creates a dependency graph. This facilitates efficient and accurate scheduling by identifying parallel sub-tasks and assessing the structural importance of each task. More importantly, we design a self-reinforced training method to boost the task allocation capability of SLM, requiring no human annotation, but only the feedback of task execution. Our training method leverages a novel tree search algorithm that accounts for both the uncertainty of language models and the task performance to optimize sub-task allocation decisions. As a result, we can create a high-quality sub-task allocation dataset without human supervision, based on which, we train a lightweight adapter that significantly boosts the SLM's ability to allocate sub-tasks. It is important to note that the adapter fully preserves the general capabilities of the SLM, as it does not modify SLM's parameters. Instead, it introduces a detachable decoding head specialized for sub-task allocation.

We conduct extensive experiments on seven widely adopted LLM agent benchmarks, covering a variety of scenarios, including logical reasoning, web browsing, math problems solving, and commonsense reasoning. The results demonstrate that our DoT framework significantly reduces LLM costs while maintaining reasoning accuracy comparable to the best baseline methods across all benchmarks. Specifically, the average reasoning time and API costs are reduced by 66.12\% and 83.57\%, respectively, compared to the most accurate baselines. Additionally, an ablation study confirms the effectiveness of our key model design choices. Moreover, the DoT framework consistently achieves superior cost-accuracy trade-offs compared to task referral methods that do not leverage ``Division-and-Allocate'' strategy, where the performance gain are particularly large at low budget setting. We further demonstrate that DoT exhibits strong generalization across benchmarks.

To summarize, our contributions are three-fold:

\begin{itemize}
    \item We present a novel Division-of-Thoughts (DoT) framework that fully exploits the synergy between locally deployed SLM and cloud-based LLM to power on-device agents with cost-effective reasoning. 

    
    \item We propose a self-reinforced training method to boost the SLM's task allocation accuracy without additional human annotation, and also preserve the general reasoning capabilities of SLM with a detachable, plug-and-play adapter.   

    \item We conduct extensive experiments on a wide range of benchmarks to evaluate the effectiveness of our DoT framework and found that DoT can significantly reduce LLM costs while maintaining high reasoning accuracy.
    
\end{itemize}

\begin{figure*}
    \centering
    \includegraphics[width=0.8\linewidth]{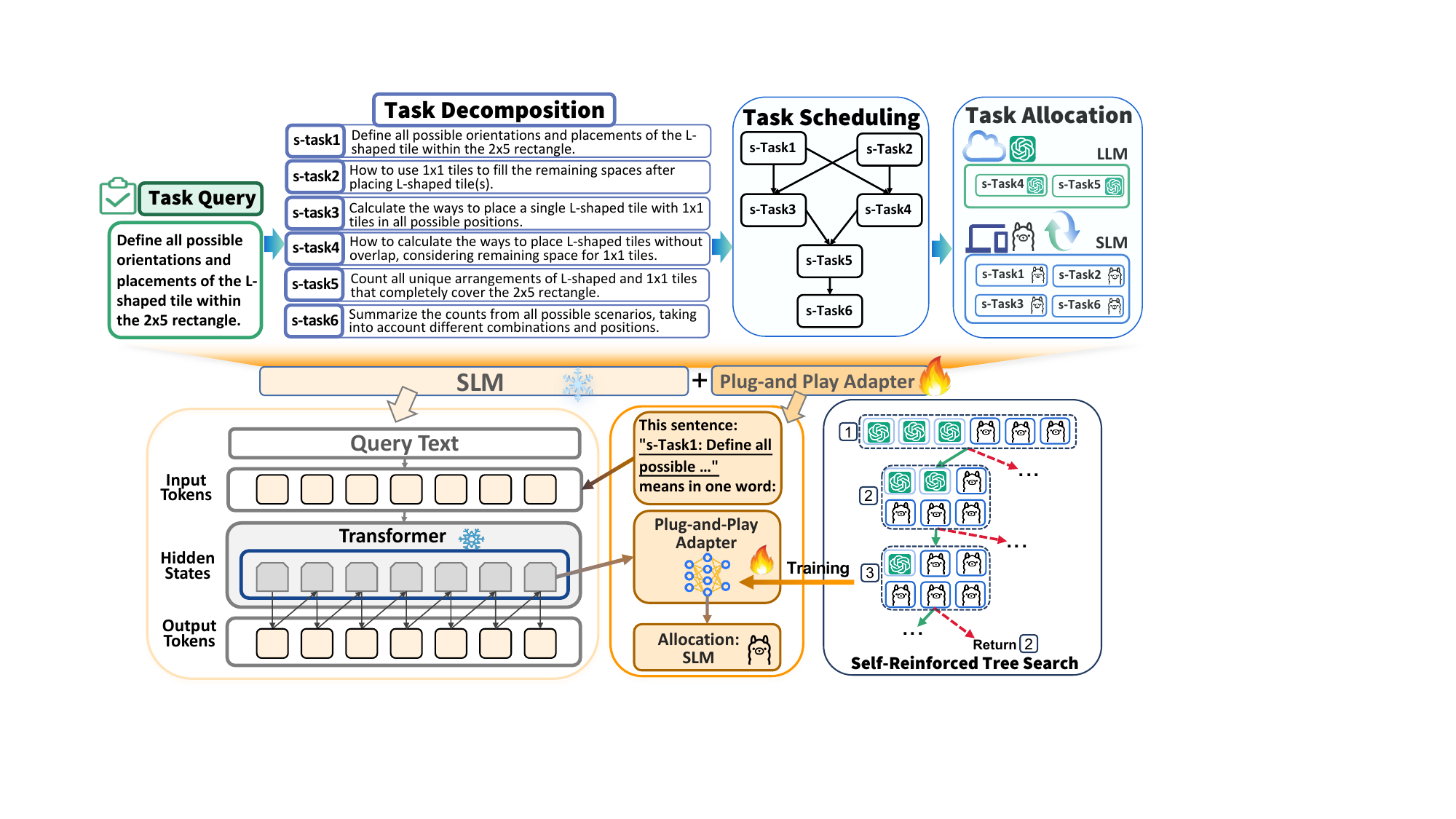}
    \vspace{-3mm}
    \caption{Overview of Our Proposed DoT Framework.}
    \label{fig:framework}
\end{figure*}
\section{Related Work}

\subsection{Reasoning with LLMs}

Recent years have seen remarkable advancements in LLMs, with increasing model scale leading to powerful emergent reasoning capabilities.~\cite{wei2022emergent, xu2025towards}. Moreover, various prompt engineering techniques have been proposed to further extend LLMs' reasoning capability.
Chain-of-Thought (CoT)~\cite{wei2022chain} improves reasoning performance by incorporating manually crafted decomposition steps into the prompt, allowing the LLM to follow the step-by-step resolution process. Building on this, Zero-shot CoT~\cite{kojima2022large} achieves similar effects by simply adding the phrase "Let's think step by step," enabling LLMs to automatically decompose and execute tasks.
From the linear structure of CoT, more complex frameworks have been introduced, such as Tree-of-Thought~\cite{yao2023tree} and Graph-of-Thought~\cite{besta2024graph}, which push the boundaries of reasoning through branching reasoning paths. However, these methods face increased resource demands and time complexity, and they heavily rely on manually predefined steps tailored to specific tasks, which significantly limits their generalizability. Additionally, there are some approaches, differing from the idea of predefined reasoning structures, that focus on how to decompose a complex problem into simpler ones for more effective resolution~\cite{ zhou2022least, khot2022decomposed, shang2024agentsquare, shang2024defint}.

Underlying these methods is the intuition of task decomposition and extensive empirical evidence has demonstrated its effectiveness. This inspires us to explore the potential of task decomposition in edge-cloud collaboration, allowing part of the task to be addressed by cloud-based models while the rest is handled by on-device models, thereby achieving a fine-grained collaborative framework.


\subsection{On-device LLM Agent}
The rise of LLMs has revolutionized AI applications, sparking interest in personal AI assistants and mobile automation tools. As on-device intelligent agents gain popularity, users expect seamless, real-time AI support on their smartphones~\cite{gong2024population}. However, the limited computational and storage capabilities of edge devices pose challenges in deploying powerful models for these agents. Under these constraints, edge-cloud collaboration is a practical solution.
Apple's latest research~\cite{gunter2024apple} exemplifies this synergy by combining an efficient on-device model, AFM-on-device, with a powerful cloud-based model, AFM-server. This approach balances device limitations with the high-performance needs of Apple’s AI features. Similarly, \cite{chen2024octo} addresses edge-device limitations by splitting task planning and execution between two models—Octo-planner and Octopus—focused on efficiency and adaptability. Both approaches highlight the trend of edge-cloud collaboration to ensure powerful, low-latency AI experiences, though Apple leans more on cloud support while Octo-planner emphasizes on-device optimization.


\section{Preliminaries}

\textbf{Problem Definition}
Denote the local deployed SLM as $\mathcal{M}_\mathcal{D}$, and the cloud-based LLM as $\mathcal{M}_\mathcal{C}$.
The user's original query is restricted to the edge model for task decomposition and allocation, while the resulting sub-tasks can be resolved by either $\mathcal{M}_\mathcal{D}$ or $\mathcal{M}_\mathcal{C}$.
The entire set of reasoning tasks is represented as $\mathcal{T} = \{T_1, T_2, \dots, T_n\}$.
Let the reasoning accuracy over the entire task set be denoted as $Acc$, with the API cost represented by $C_{Api}$, and the completion time denoted as $C_{Time}$. 
For each task $T$, denote the decomposition process as:
\begin{equation}
T\rightarrow \{t^1,t^2,...t^{k}\}
\end{equation}
Based on the decomposed subtasks $t^i$, the model allocation scheme can be denoted as:
\begin{equation}
M :t^i\mapsto\{ \mathcal{M}_\mathcal{D} , \mathcal{M}_\mathcal{C}\}
\end{equation}
which prioritizes assigning simple subtasks to on-device SLM, while invoking the cloud-based LLM for handling complex subtasks.

The goal of our optimization is to minimize the discrepancy between the model's allocation scheme $M$ and the optimal scheme $M^*$:
\begin{equation}
   \min|M - M^*| 
\end{equation}
The optimal scheme $M^*$ is derived through a search strategy that maximizes SLM usage while maintaining accuracy. During the optimization process, as the allocation scheme gradually approaches the optimal solution, both time cost $C_{Time}$ and API cost $C_{Api}$ decrease, while $Acc$ remains well-maintained.

\section{Methodology}


\subsection{Division-of-Thoughts Framework}
The \textit{division-of-labour} is a prominent concept in economics, famously introduced by Adam Smith. It refers to the separation and allocation of tasks within any economic system or organization, enabling participants to specialize in specific areas based on their unique capabilities. This specialization allows individuals, organizations, and nations to optimize productivity by leveraging specialized skills, equipment, and resources. This concept inspires us to consider a similar approach in the context of edge-cloud collaboration, where user queries can be intricately divided: simpler sub-tasks can be assigned to SLMs while more complex ones are allocated to LLMs. This sub-task-level division of labor is expected to reduce reasoning costs while maintaining reasoning performance, enabling more efficient collaboration. 

\begin{figure}
    \centering
    \includegraphics[width=\linewidth]{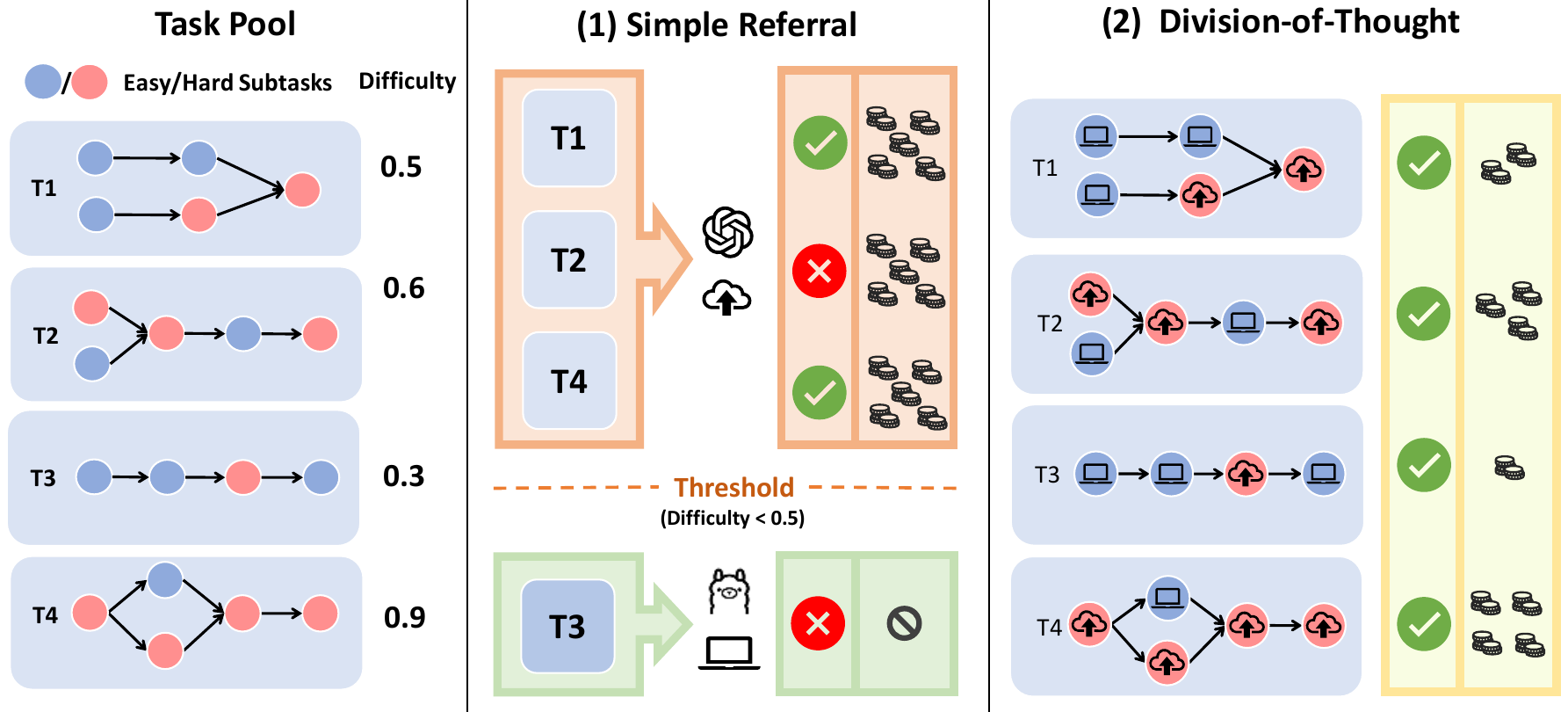}
    \vspace{-6mm}
    \caption{Advantage of ``Division-and-Allocate'' Strategy}
    \label{fig:compare}
    \vspace{-5mm}
\end{figure}

Building on this intuition, we develop the \textbf{DoT} edge-cloud collaboration framework. As illustrated in Figure \ref{fig:framework}, our framework is divided into three components. The first component is the \textit{Task Decomposer}, which breaks down the user's query into several simpler and independent sub-tasks.
The second component is the \textit{Task Scheduler}, which is leveraged to determine the pairwise dependencies between sub-tasks and constructing a task dependency graph based on these dependencies.
The third component is the \textit{plug-and-play LLM adapter}, responsible for assigning each sub-task to the appropriate models.
The adapter extracts sentence embeddings from the SLM and maps them to difficulty coefficients, which serve as the basis for the model allocation for sub-tasks. Importantly, the training of the adapter does not require modifying the LLM's parameters, ensuring that the LLM's question-answering remains unaffected.
Once the appropriate models have been assigned to each sub-task, reasoning proceeds along the order defined by the constructed dependency graph, leading to the final results.
Figure \ref{fig:compare} clearly demonstrates that, compared to the simple referral strategy, DoT exhibits significant advantages in both accuracy and efficiency.

\subsection{Decomposing User Query}
As emphasized in the \textit{division-of-labour}, an effective division method serves as the foundation for collaborative work. The granularity and accuracy of division directly influence the quality and efficiency of the collaboration. 
Prior research on model collaboration has primarily focused on query-level granularity and relied heavily on manual task decomposition strategies. A typical example is the ToT approach in Game-of-24, which employs a fixed two-step process of proposal generation and evaluation. This rigid, manually-designed workflow lacks generalizability across diverse tasks, underscoring the necessity for automated, flexible, and fine-grained task decomposition methods.

To address these challenges, we develop a fine-grained task decomposition method based on the powerful \textit{In-Context Learning (ICL)} capabilities of LLMs. We leverage a meta-prompt that incorporates ``chain-of-thought''-like prompting with hand-crafted task decomposing examples, to elicit the inherent \emph{planning} abilities in SLMs. 
For each benchmark, We randomly selected 8 samples, manually performed step-by-step task decomposition, and incorporated these steps into the prompts.
To enable LLMs to independently solve each sub-task and effectively use the answers from preceding tasks as references for subsequent reasoning, we place great emphasis on the independence and clarity of the decomposed sub-tasks and have incorporated targeted cues in the prompts design.

\subsection{Task Scheduling via Dependency Graph}
\label{method:dep}
\begin{figure}
    \centering
    \includegraphics[width=0.96\linewidth]{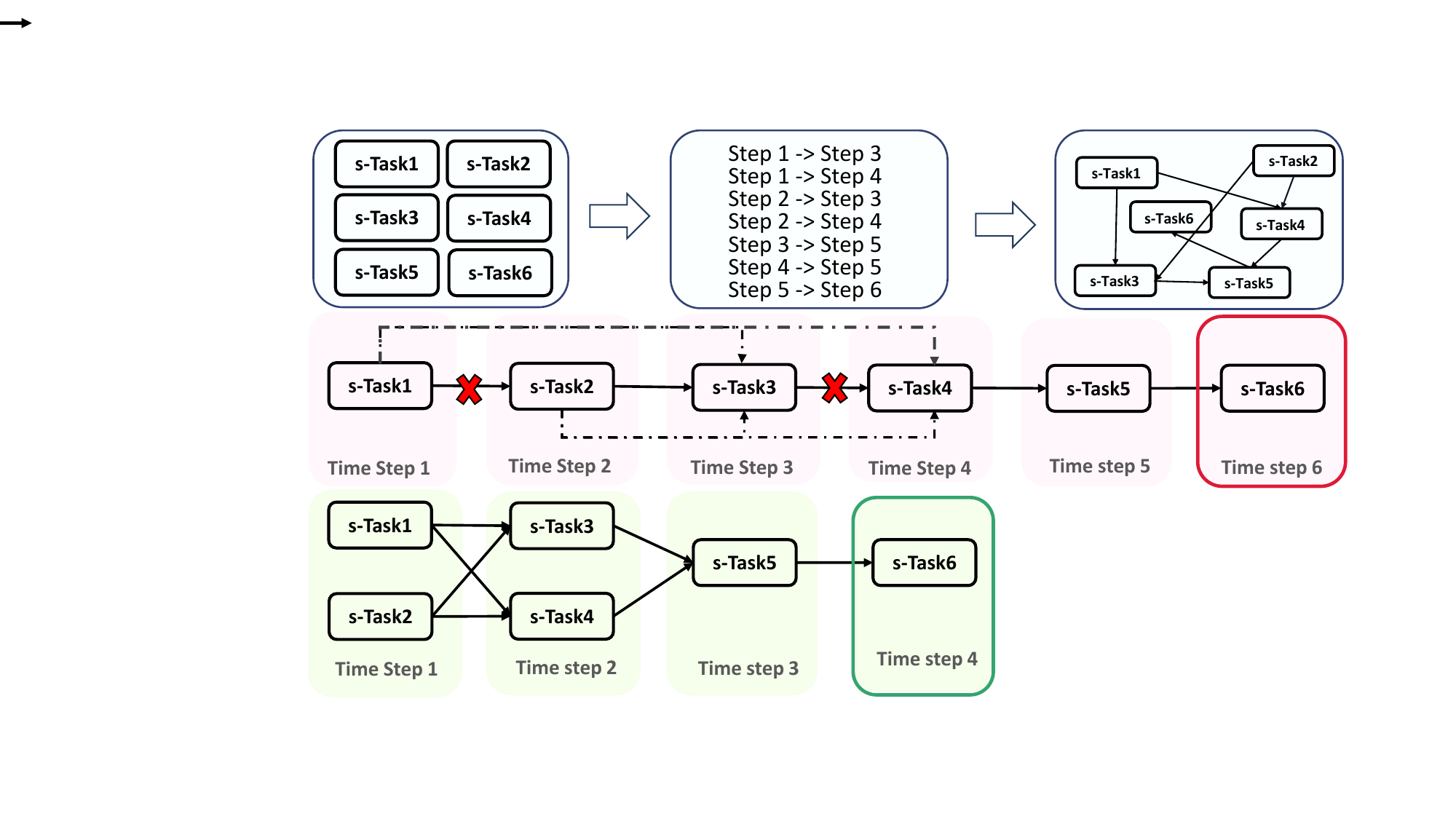}
    \vspace{-3mm}
    \caption{Illustrating Dependency Graph of Task Scheduling}
    \label{method:flow-Compare}
    \vspace{-5mm}
\end{figure}

We employ a three-step approach to schedule all the sub-tasks effectively:
\textbf{Dependency Judgement.} Prompting LLM agent to assess pairwise dependencies between sub-tasks. The prompts are as follows: \textit{"Please list the dependencies in the format 'Subproblem A [xxx] -> Subproblem B [xxx]' indicating that Subproblem A must be completed before Subproblem B can start."}
\textbf{Graph Construction.} Converting dependencies into a dependency graph. Based on the clear directional dependencies, constructing the graph is straightforward. However, we further trace back from the final sub-task node to calculate the depth of each sub-task node in the graph (analogous to the height in a tree structure). Sub-tasks at the same depth are independent of each other and can be inferred in parallel. Depth serves as the inference batch, with batches processed sequentially while tasks within a batch are reasoned in parallel. As shown in Figure \ref{method:flow-Compare}, compared to sequentially reasoning through all sub-tasks, our graph structure can capture more precise inferential relationships while incurring fewer time costs.
\textbf{On-Graph Reasoning}. We start by solving the sub-tasks from the shallowest depth batch, running the sub-tasks within the same batch in parallel. After completing one batch, we proceed to the next. During the reasoning of a specific sub-task, only the results of its prerequisite sub-tasks, rather than all previously solved tasks, are included in the prompts, which enables efficient graph-based reasoning.

\subsection{Task Allocation with Reinforced SLM}
\label{method:allocation}

For the decomposed sub-tasks, we aim to assess their difficulty based on the task descriptions and allocate either cloud-based or edge-side models for execution. Using LLMs to evaluate difficulty introduces additional inference costs and often fails to accurately assess the sub-task’s complexity, which motivates us to train a model specifically designed for task allocation with sentence embeddings.
\textit{Sentence embeddings}, which map a sentence to a fixed-size vector capturing its semantic meaning and context, have seen extensive application in natural language processing for their lightweight accessibility and the strong ability to capture sentence semantics. We can obtain a sentence embedding for each sub-task's prompt and then map the embedding to the corresponding difficulty. Given that LLMs are already deployed on the edge side and have been shown to serve as effective sentence embedders, we plan to leverage the local deployed SLM to produce sentence embeddings.

However, autoregressive LLMs lack specialized tokens such as BERT’s [MASK] or [CLS], which are typically used in transformer-based models for embedding tasks. The exploration of Ting Jiang et al.~\cite{jiang2023scaling} helps us overcome this limitation. 
Prompt-based method is refined specifically for autoregressive LLMs by instructing the model to generate the next word that captures the semantic meaning of the input sentence. Specifically, a simple but effective template can be constructed for each piece of text: \textit{This sentence: "[text]" means in one word: "}, where \textit{[text]} represents the input sentence, and the LLM is prompted to generate the next token that encapsulates the core meaning of the sentence in a single word. The last generated token plays a critical role, as we extract its hidden state from the model and use it as the sentence embedding.

\textbf{Boosting SLM with Plug-and-Play Adapter.}
Despite the varying lengths of sub-task descriptions, the sentence embeddings maintain a consistent dimensionality. For efficiency, we append a multi-layer perceptron (MLP) to the transformer module of the LLM to map the embeddings to a difficulty score. Given that both the edge side and the cloud host only one model, we can simply set the scores as 0 (simple, for SLM) or 1 (complex, for LLM). The output from the MLP is translated into a model designation and passed to the agent handling the current reasoning step, providing a concrete allocation strategy for edge-cloud collaboration. 
Moreover, the adapter requires no modifications to SLM's parameters. It essentially serves as a flexible and extensible adapter for the SLM, avoiding the need to store two sets of parameters on the device.

\begin{figure}
    \centering
    \includegraphics[width=1.01\linewidth]{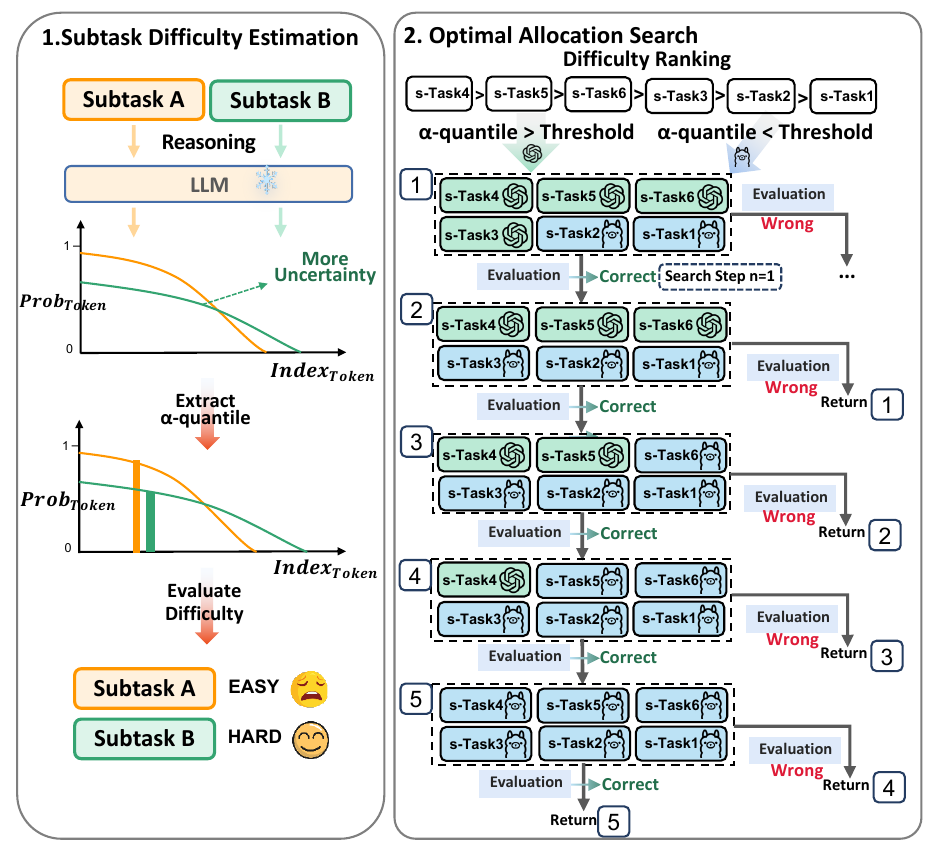}
    \vspace{-6mm}
    \caption{Tree Search-Based Dataset Construction Process}
    \label{method:dataConstruction}
    \vspace{-4mm}
\end{figure}

\textbf{Creating Training Data with \textbf{$\alpha$}-Tree Algorithm.}
How cam we construct a dataset to train our adapter?
We develop an efficient allocation optimization strategy, $\alpha-Tree$,  which consists of two key components: the sub-task difficulty ranking based on the $\alpha$-quantile token probability, and a tree-based search strategy guided by the ranking, allowing us to generate a large-scale optimal allocation dataset with low cost and high speed.
$\alpha$-quantile refers to calculating a specific quantile from the token probabilities generated by LLM during inference. Unlike traditional methods that aggregate probabilities through summation or averaging, $\alpha$-quantile focuses on a specific portion of the uncertainty distribution, such as the minimum ($\alpha$ = 0) or a higher percentile (e.g., $\alpha$ = 0.8), which provides a fine-grained measure of uncertainty.
Denote the input context as $x$, and the output tokens as $\hat{y}_i'$. The $\alpha$-quantile value is denoted as:
\begin{equation}
\begin{aligned}
s_{\text{quant}}(x, \alpha) &\doteq \text{quantile}_{\alpha}\left(p^{(1)}(\hat{y}_1' \mid x), \right.  p(\hat{y}_2' \mid x, \hat{y}_1'), \dots, \\
& \left. \quad p(\hat{y}_n' \mid x, \hat{y}_1', \dots, \hat{y}_{n-1}')\right)
\end{aligned}
\end{equation}

\begin{algorithm}
	\caption{Self-reinforced dataset construction method: $\alpha-Tree$}
	\begin{algorithmic}[1]
         \Require  
            Tasks which has been decomposed into several subtask: 
            $\left\{T_1=\{t^1_1,..,t^{k_1}_1\}, T_2=\{t^1_2,..,t^{k_2}_2\},...,T_n=\{t^1_n,..,t^{k_n}_n\}\right\}$ 
        \Ensure  
            Fine-grained device-cloud model allocation scheme applied to each subtask within every task
            ($m_i^j\in \{\mathcal{M}_\mathcal{D},\mathcal{M}_\mathcal{C}\}$):
            $\left\{M_1=\{m^1_1,..,m^{k_1}_1\}, M_2=\{m^1_2,..,m^{k_2}_2\},...,M_n=\{m^1_n,..,m^{k_n}_n\}\right\}$

		\For {task id i}
    		\State Perform on-graph reasoning for eash subtask: $Answer = \{a^1_i,..,a^{k_i}_i\}$ with $Prob_{token} = \{p^1_i,..,p^{k_i}_i\}$ 
                \State Calculate the $\alpha$-quantile value for each token sequence: $V_{\alpha-quantile} = \{v^1_i, v^2_i,...,v^{k_i}_i\}$
                \State Set a threshold $\theta$ to obtain the initial allocation strategy: $M_i=\{m^1_i,..,m^{k_1}_i\}$ where
                    $ m^j_i =
                    \begin{cases} 
                    \mathcal{M}_\mathcal{D} & \text{if } v^j_i>\theta \\ 
                    \mathcal{M}_\mathcal{C} & \text{if } v^j_i<\theta 
                    \end{cases}
                    $
                \State Reasoning on the task to obtain: \textit{result = True or False}
                
                \While{True}
            		\If {\textit{result == True}}
            		      \State Find all subtasks assigned to $\mathcal{M}_\mathcal{C}$, $M_{Ci} = \{ m \in M_{i} \mid m == \mathcal{M}_\mathcal{C} \}$, select N subtasks with the highest $\alpha$-quantile probabilities, and reassign them to $\mathcal{D}$. Get new $M_{i}^\prime$.
                        \Else
                            \State Find all subtasks assigned to $\mathcal{M}_\mathcal{D}$, $M_{Di} = \{ m \in M_{i} \mid m == \mathcal{M}_\mathcal{D} \}$, select N subtasks with the lowest $\alpha$-quantile probabilities, and reassign them to $\mathcal{M}_\mathcal{C}$. Get new $M_{i}^\prime$.
            		\EndIf
                        \State Reasoning to obtain new result $result^\prime$ 
                        \If{$len(M_{Di})==0$ or $len(M_{Ci})==0$ or $result^\prime \neq result$}    
                            \State break 
                        \EndIf
                        \State $M_{i} = M_{i}^\prime$
    		\EndWhile
		\EndFor
	\end{algorithmic} 
\label{algorithm}
\end{algorithm}

In our implementation, we employ the SLM to answer all decomposed sub-questions, recording the sampling probability corresponding to each token. We then apply a consistent alpha value to extract the $\alpha$-quantile from each probability sequence and ranked the $\alpha$-quantile values for all sub-task responses. Since higher task difficulty leads to greater uncertainty in the model’s answers, resulting in lower token sampling probabilities, we reverse the $\alpha$-quantile ranking to obtain the difficulty order of the sub-tasks.

Based on the $\alpha$-quantile values of the sub-tasks, we set a fixed allocation threshold: sub-tasks with values exceeding this threshold are assigned to SLM, while those below the threshold are assigned to LLM, forming the initial model allocation. From this initial allocation, we perform collaborative reasoning with SLM and LLM. If the reasoning result is correct, we reassign $n$ of the sub-tasks currently handled by LLM, specifically those with the highest probability of being correctly answered, to SLM. Conversely, if the answer is incorrect, we transfer $n$ of the sub-tasks handled by SLM, specifically those with the lowest probability of being misanswered, to LLM. 
As shown in Figure \ref{method:dataConstruction}, the structure of the search exhibits a tree-like form.
The algorithmic workflow for dataset construction is illustrated in Algorithm \ref{algorithm}.

\vspace{-2mm}
\section{Experiments}
\begin{table*}
\centering
\renewcommand\arraystretch{1}
\setlength{\tabcolsep}{0.5mm}
\resizebox{\textwidth}{!}
{
\begin{tabular}{c|ccc|ccc|ccc|ccc|ccc|ccc|ccc}
\multirow{3}{*}{\textbf{Model}} & \multicolumn{6}{c|}{\textbf{Logical Reasoning}}                                                                                       & \multicolumn{3}{c|}{\textbf{Web Browsing}}                      & \multicolumn{9}{c|}{\textbf{Solving Math Problems}}                                                                                                                                             & \multicolumn{3}{c}{\begin{tabular}[c]{@{}c@{}}\textbf{Commonsense}\\\textbf{Reasoning}\end{tabular}}  \\ 
\cline{2-22}
                       & \multicolumn{3}{c|}{\textbf{P3}}                                  & \multicolumn{3}{c|}{\textbf{SCAN}}                                & \multicolumn{3}{c|}{\textbf{WebShop}}                           & \multicolumn{3}{c|}{\textbf{MATH}}                                & \multicolumn{3}{c|}{\textbf{CHAMP}}                          & \multicolumn{3}{c|}{\textbf{DROP}}                           & \multicolumn{3}{c}{\textbf{CSQA}}                                                                     \\ 
\cline{2-22}
                       & $Acc$                 & $C_{Time}$            & $C_{API}$             & $Acc$                 & $C_{Time}$            & $C_{API}$             & \textit{Acc}          & $C_{Time}$            & $C_{API}$           & $Acc$                 & $C_{Time}$            & $C_{API}$             & $Acc$            & $C_{Time}$            & $C_{API}$             & $Acc$            & $C_{Time}$            & $C_{API}$             & $Acc$         & $C_{Time}$            & $C_{API}$                                                         \\ 
\hline
COT (GPT-4o)           & \textbf{\underline{42\%}}  & \underline{35.8} & \underline{4.45\textcent}  & \textbf{\underline{68\%}}   & \underline{9.21}  & \underline{2.75\textcent}   &     35\% &      30.9  &    10.65\textcent                       & 51.5\% & 34.5 & 5.34\textcent  & 55.5\% & 26.4  & 4.45\textcent  & 80\%   & 11.6   & 1.30\textcent   & 80\%   &  17.0  & 3.60\textcent        \\
TOT (GPT-4o)           & 38\%  & 93.1 & 14.55\textcent & 52\%   & 32.5 & 9.82\textcent &  \textbf{\underline{36\%}}  &  \underline{62.4}    & \underline{47.34\textcent}                              & \textbf{\underline{63\%}}   & \underline{60.5} & \underline{9.97\textcent}  & \underline{57\%}   & \underline{64.2}  & \underline{11.65\textcent}  & \underline{80.5\%} & \underline{40.2} & \underline{5.41\textcent}  & \underline{82\%}   & \underline{98.8} & \underline{20.50\textcent}         \\
COT (Llama 3-8B)        & 5.5\% & 18.1 & N/A  & 17\%   & 5.0  & N/A   &   0.0\%  &   10.5   &  N/A                             & 10\%   & 21.1  & N/A   & 19\% & 13.1 & N/A   & 72\%   & 3.8  & N/A   & 70\%   & 8.4 & N/A          \\
TOT (Llama 3-8B)        & 5.5\% & 58.3 & N/A  & 13\%   & 21.8 & N/A   &   1.4\%  &   22.5  &  N/A                             & 29.5\% & 49.0 & N/A   & 25\%   & 68.1 & N/A   & 65\%   & 27.8 & N/A   & 68.5\% & 89.4 & N/A          \\
DataShunt              & 14\%  & 25.1 & 2.45\textcent  & 23.5\% & 7.6  & 1.72\textcent  &   34\%  &  30.9   & 8.35\textcent                            & 16\%   & 24.9 & 1.66\textcent & 34\%   & 19.1 & 2.98\textcent  & 74\%   & 8.6  & 0.60\textcent & 73\%   & 10.4  & 1.28\textcent        \\ 
DoT (ours)             & 41\%  & 23.5 & 1.58\textcent & 63\%   & 5.5  & 1.20\textcent  &  31\%   &   17.2   &     4.97\textcent                      & 59\% & 22.6 & 1.02\textcent   & \textbf{58\%}   &  16.1   & 0.84\textcent   & \textbf{85\%}   & 4.9 & 0.32\textcent  &   \textbf{82\%}   & 9.9  & 0.49\textcent    \\     
Improvement             &  $\downarrow$2.38\%   &    $\downarrow$34.36\%      &    $\downarrow$64.49\%  &  $\downarrow$7.35\%  & $\downarrow$40.28\% & $\downarrow$56.36\%  & $\downarrow$13.89\% & $\downarrow$72.43\% & $\downarrow$89.95 & $\downarrow$6.35\%  &  $\downarrow$62.72\%              &     $\downarrow$89.50\%     &    $\uparrow$1.75\%   &  $\downarrow$74.92\%        &       $\downarrow$92.79\%              &  $\uparrow$5.59\%      & $\downarrow$87.81\%               &      $\downarrow$94.09\%    &    0\%    &      $\downarrow$89.98\%          &    
$\downarrow$97.60\%                                           
\end{tabular}
}
\vspace{2mm}
\caption{Performance of DoT and baselines on 7 benchmarks. 
$\textbf{C}_{\textbf{Time}}$ and $\textbf{C}_{\textbf{API}}$ are averaged expense for each task, where time consumption is measured in seconds, and API cost is measured in US dollar cents (\textcent). \textit{N/A} appears in experiments where reasoning is conducted solely using LlaMA without invoking the OpenAI's API key. In each benchmark, the highest reasoning accuracy is highlighted in bold. The results of the baseline with the highest \textit{Acc} are underlined which will be used to compute the "Improvement" in the last row.}
\label{tbl:mainResult}
\vspace{-7mm}
\end{table*}

\subsection{Experimental Setup}

\textbf{Benchmarks.}
We validate the effectiveness of our framework across seven open-source benchmarks.
For each benchmark, we randomly select 8 instances for manual annotation to serve as in-context learning examples of task decomposition. Additionally, we randomly extract 200 questions from each benchmark to constitute our test set. The benchmarks are categorized into four groups:
\begin{itemize}
    \item \textbf{Logical Reasoning.} 
    We select P3~\cite{schuster2021programming} and SCAN~\cite{lake2018generalization} for evaluation.
    These benchmarks place a stronger emphasis on logic, involving challenges such as traversal, backward reasoning, and anomaly detection, which require a higher level of logical coherence and rigor.
    \item \textbf{Math Problems Solving.} This category of benchmarks primarily addresses mathematical problems, involving computation, mathematical knowledge, and problem-solving techniques. It is widely regarded as a crucial test of LLMs' reasoning abilities. We select three benchmarks for this category: MATH~\cite{hendrycks2021measuring}, CHAMP~\cite{mao2024champ}, and DROP~\cite{dua2019drop}.
    \item \textbf{Commonsense Reasoning.}  CSQA~\cite{talmor2018commonsenseqa} is a widely used commonsense reasoning dataset that places less emphasis on reasoning capabilities but requires a broader knowledge base. The difference in LLMs' parameter scales leads to disparities in the knowledge systems they possess, making CSQA well-suited for evaluating our collaborative reasoning scenario.
    \item \textbf{Web Browsing.} We choose WebShop~\cite{yao2022webshop} which tests LLMs as interactive agents in real-world scenarios. It challenges LLM agents to navigate web pages, interpret complex queries, and take appropriate actions to fulfill user requests.
    This benchmark closely aligns with AI assistant scenarios, encompassing  instructions understanding, query reformulation, and strategic information exploration.
\end{itemize}

\textbf{Baselines and Prompting Methods.} 
Given the constraints of the edge-cloud collaborative setting, we establish three baselines.

\begin{itemize}
    \item \textbf{CoT~\cite{wei2022chain}.} This baseline decomposes the problem into sub-steps, solving each sequentially. We use a single LLM throughout, executing each step only once without iteration.
    \item \textbf{ToT~\cite{yao2023tree}.} ToT explores multiple solutions (M=5) for each sub-step. We use a scoring mechanism to retain the N=3 top-scored paths, with the highest-scored path determining the final answer. ToT also employs a single LLM for reasoning.
    \item \textbf{DataShunt~\cite{chen2024data}.} This approach dynamically selects between on-device SLM and cloud-based LLM at the start of each task. It prompts the LLM to evaluate task complexity, assigning a \textit{solving model} that is then used consistently for all sub-steps.
\end{itemize}
In the task decomposition processes involved in CoT and ToT, we employed the same prompting method used in our DoT, applying eight hand-crafted examples to each benchmark. These examples are randomly sampled from the dataset and are orthogonal to the test set. The specific prompts can be found in Appendix \ref{appendix:imple_details}.

\textbf{Selection and Deployment of LLMs}
We used GPT-4o~\cite{GPT-4o} as cloud-based LLM and Llama 3-8B~\cite{LLaMA3} as on-device LLM as its parameter scale is theoretically deployable on edge devices.
The Llama 3-8B model is deployed on a local A100 GPU with a context length limit of 8192. The parameter count of our adapter is only 13,109,249, approximately 1/800th of the parameters in local Llama.

\textbf{Evaluation.} 
For evaluation, we establish three metrics: $Acc$, $C_{time}$, $C_{API}$. Six of the benchmarks, excluding WebShop, have deterministic results. For instance, the results for CSQA are single-letter options, allowing direct determination of correctness. Similarly, for P3, the reasoning output can be passed into a problem function, and if the function returns True, the reasoning is deemed correct.
For WebShop, we use the evaluation environment provided by the original benchmark to average the reward of all purchased items on the level of satisfaction according to the constraints.


\vspace{-3mm}
\subsection{Main Results}

Comparison between our method and baselines is shown in Table 1, we have highlighted in bold the highest accuracy results among the five baseline experiments on each benchmark, while the associated costs are underlined. We compute the relative improvement of our results compared to the baseline with the highest accuracy.

The experimental results demonstrate that our approach significantly reduces costs while maintaining accuracy within an acceptable range of decline. Across seven benchmarks, the relative changes in accuracy compared to the best baseline results are: -2.38\%, -7.35\%, -13.89\%, -6.35\%, 1.75\%, 5.59\%, and 0\%. Notably, positive improvements are observed in some cases, suggesting that our framework has the potential to enhance LLMs' reasoning capability. This enhancement is further discussed in ablation study~\ref{exp:ablation}. At the same time, our approach achieves a substantial reduction in cost compared to the baseline with the highest accuracy, with an average time reduction of 66.12\% and an average API cost reduction of 83.57\% which far exceeded the decline in accuracy.

Moreover, on P3 and SCAN benchmarks, CoT achieves the best performance, indicating that a sequential linear reasoning structure is better suited for these task types. In contrast, on more complex mathematical reasoning benchmarks like MATH, CHAMP, and WebShop, ToT achieves the highest reasoning accuracy, demonstrating the effectiveness of the multiple-proposal strategy. However, our approach incurs less than half of CoT's cost and only one-tenth of ToT's cost while still achieving comparable performance, highlighting the broad applicability of our graph-structured reasoning and the effectiveness of the edge-cloud collaboration. 

\begin{figure}[h]
    \centering
    \subfigbottomskip = -3pt
    \subfigure{
        \includegraphics[width=0.9\linewidth]{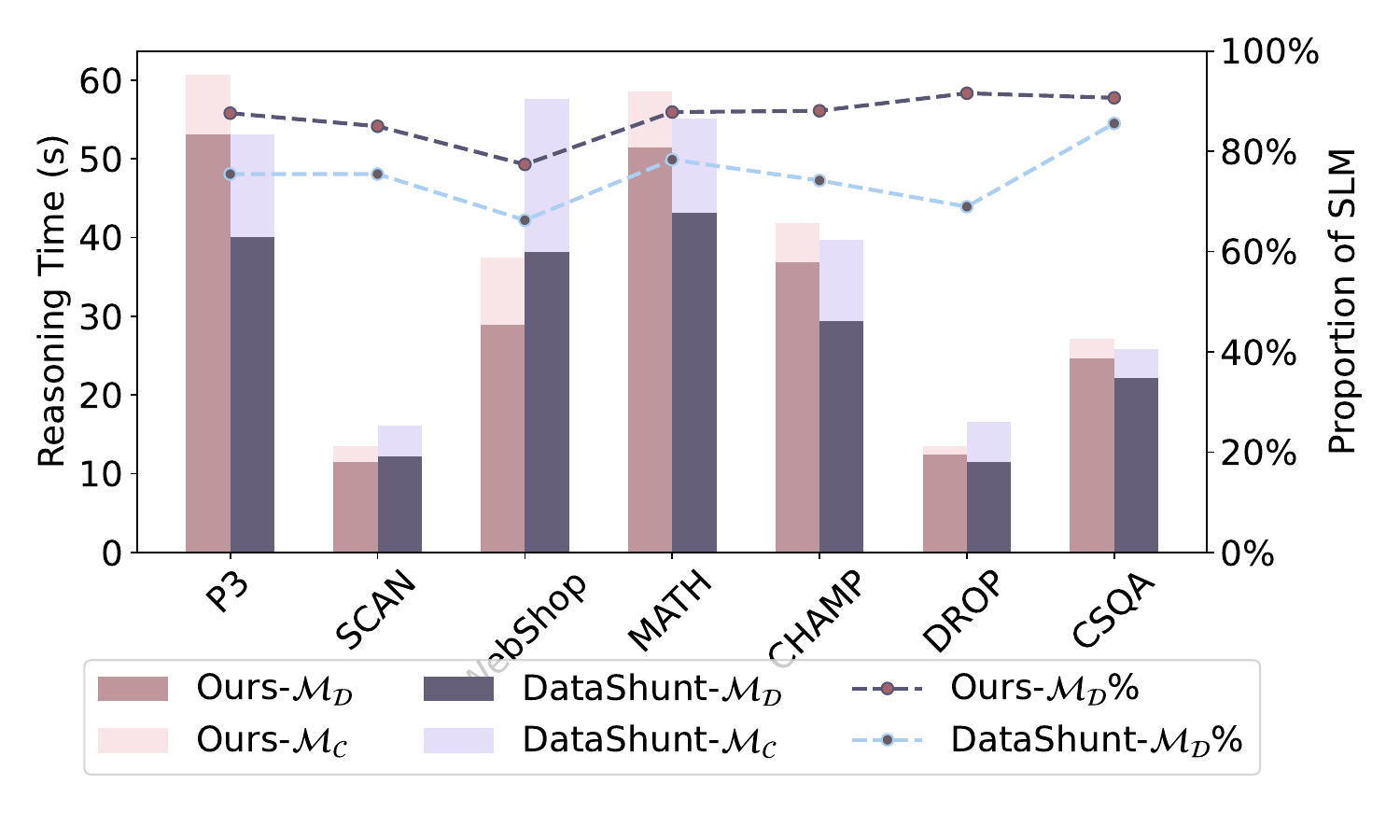}
    }
    \quad
    \subfigure{
        \includegraphics[width=0.9\linewidth]{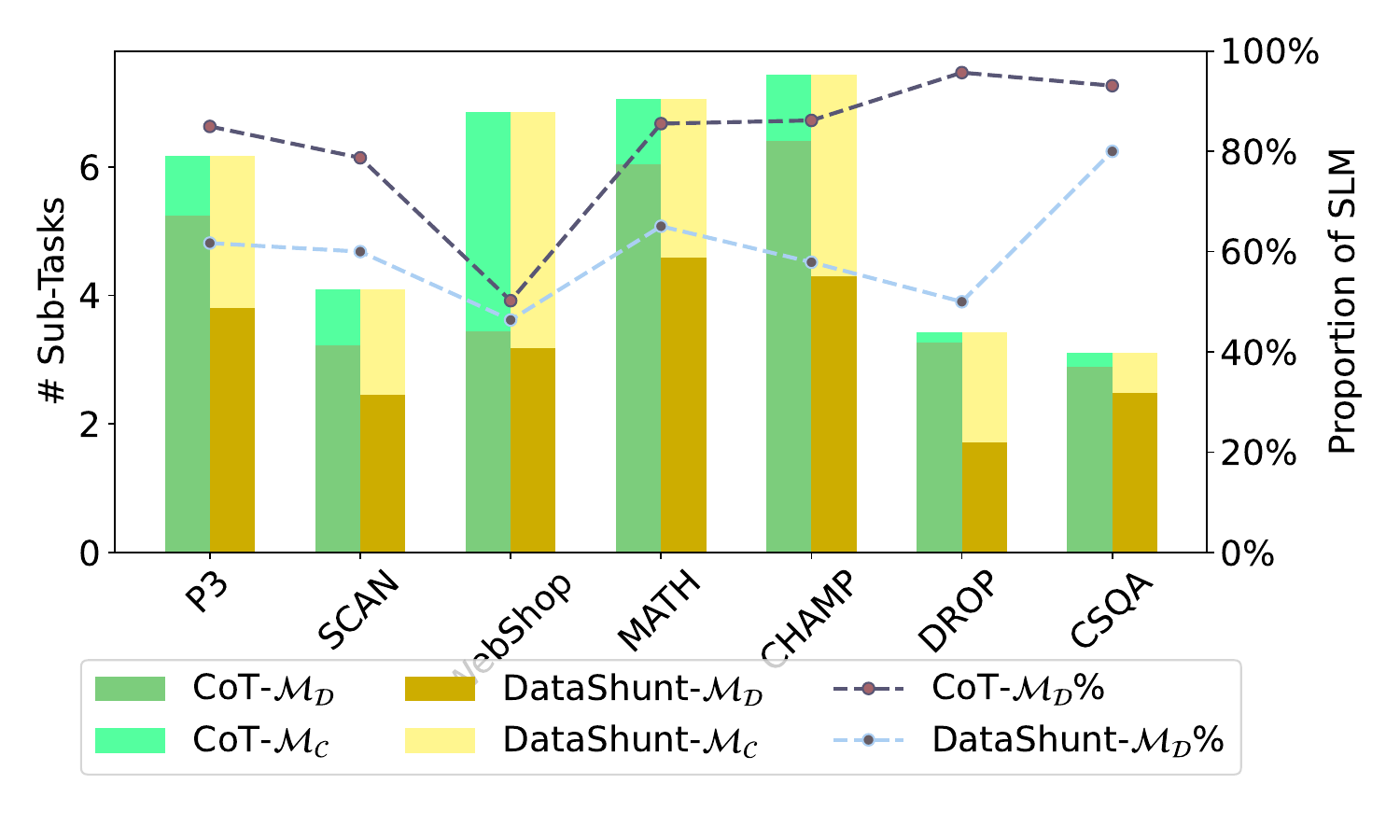}
    }
    \vspace{-5mm} 
    \caption{Proportion of SLMs in time cost and \# sub-tasks.}
    \label{exp: SLM}
    \vspace{-4mm}
\end{figure}

To evaluate whether our collaborative approach fully leverages locally deployed SLMs for reasoning, we calculate the proportion of reasoning time by the SLMs relative to the overall reasoning time cost, as well as the percentage of sub-tasks assigned to SLMs. These metrics were compared against the baseline \textit{DataShunt}, and the results are presented in Figure \ref{exp: SLM}. 
The stacked bar chart illustrates the specific values of SLM and LLM, while the line graph reflects the proportion of SLM.
As shown in the figure, across all benchmarks, our allocation strategy makes more comprehensive use of SLMs, with the reasoning time and task allocation percentages exceeding those of the baseline by an average of 11.99\% and 21.92\%, respectively.
\vspace{-5mm}

\begin{table}
\centering
\resizebox{\linewidth}{!}
{
\begin{tabular}{c|c|c|c|c}
\textbf{Methods}              & \textbf{SLM Ratio} & \textbf{SR} & \textbf{\# Evaluation} & \textbf{API Cost}  \\ 
\hline
Zero-Shot LLM                    & 53.11\%            & 92.78\%     & 1                    & \$2.56             \\
Binary-Search                & 69.46\%            & 88.45\%     & 8.456                & \$16.45            \\
$\alpha$-tree (n=1)~ ~ & 85.53\%            & 99.44\%     & 3.4589               & \$7.34             \\
$\alpha$-tree (n=2)~ ~ & 86.45\%            & 96.34\%      & 2.3445               & \$5.12            
\end{tabular}
}

\vspace{1mm}
\caption{Methods for Searching Optimal Allocation Scheme}
\label{tbl:allocation}
\vspace{-7mm}
\end{table}

\begin{table*}[htb]
\centering
\renewcommand\arraystretch{1}
\setlength{\tabcolsep}{0.5mm}
\resizebox{\textwidth}{!}{
\begin{tabular}{c|ccc|ccc|ccc|ccc|ccc|ccc|ccc}
\multirow{3}{*}{\textbf{Model}}          & \multicolumn{6}{c|}{\textbf{Logical Reasoning}}                       & \multicolumn{3}{c|}{\textbf{Web Browsing}} & \multicolumn{9}{c|}{\textbf{Solving Math Problems}}                                                           & \multicolumn{3}{c}{\begin{tabular}[c]{@{}c@{}}\textbf{Commonsense}\\\textbf{Reasoning}\end{tabular}}  \\ 
\cline{2-22}
                                & \multicolumn{3}{c|}{\textbf{P3}} & \multicolumn{3}{c|}{\textbf{SCAN}} & \multicolumn{3}{c|}{\textbf{WebShop}}      & \multicolumn{3}{c|}{\textbf{MATH}} & \multicolumn{3}{c|}{\textbf{CHAMP}} & \multicolumn{3}{c|}{\textbf{DROP}} & \multicolumn{3}{c}{\textbf{CSQA}}                                                                     \\ 
\cline{2-22}
                                & Acc  & Time & Api                & Acc  & Time & Api                  & Acc    & Time & Api                        & Acc  & Time & Api                  & Acc  & Time & Api                   & Acc  & Time & Api                  & Acc  & Time & Api                                                                                     \\ 
\hline
DoT~w/o Graph                 &    38\%  &   36.4           &   1.44\textcent     &  57\%    &   5.2          &  1.18\textcent      &  31\%   &   20.7  &     4.97\textcent    & 56\%  &  20.9  &  0.99\textcent                            & 52\%       &   27.7          &    0.82\textcent     &  80\%    &  4.2        &   0.26\textcent         &  77\%    &    9.6       &  0.48\textcent                      \\ 
DoT~w/o Allocation (GPT-4o)   &  43\%    &  39.8  &   4.80\textcent       &       64\%    &   10.7    &  3.66\textcent     & 39.5\% & 19.4 &  13.89\textcent     &    78\%         &  36.9   & 7.68\textcent    &  64\%                             &  30.4      &  5.05\textcent          &    84\%       &   12.8   &   1.69\textcent          &  85\%            &  20.6   & 1.49\textcent           \\
DoT~w/o Allocation (Llama 3-8B)   &   29\%   &    21.5         & N/A          & 52\%     &  4.2        &  N/A    &  3.5\% &  10.7  &  N/A     & 49\%    & 19.8    &  N/A                             &   47\%     &  15.8           &  N/A         &  77\%    &  4.3             &  N/A            &   76\%   &  8.9           &   N/A           \\ 

DoT~w/o Allocation (Simple Referral)   &  33\%    & 31.1  &   3.21\textcent       &       57\%    &   6.3    &  2.41\textcent     & 12.5\% &  13.5 &  6.27\textcent     &    51\%         &  26.3   & 3.54\textcent    &  53.5\%                             & 19.6     &  2.15\textcent          &    80\%       &  6.4 &   0.87\textcent          & 79\%            &  14.6  & 0.64\textcent           \\

DoT (ours)             & 41\%  &   23.5   & 1.58\textcent  & 63\%   & 5.5  &  1.20\textcent  &  31\%   &  17.2   &     4.97\textcent                          & 59\% &   22.6   & 1.02\textcent  & 58\%   & 16.1   & 0.84\textcent   & 85\%   &  4.9  & 0.32\textcent  & 82\%  & 9.9  & 0.49\textcent                                                                         
\end{tabular}
}
\vspace{1mm}
\caption{Result of ablation study.}
\label{tbl:ablation}
\vspace{-4mm}
\end{table*}

\begin{figure*}[h]
    \centering

    \parbox[b]{0.47\linewidth}{
        \centering
        \subfigure[MATH]{\includegraphics[width=0.49\linewidth]{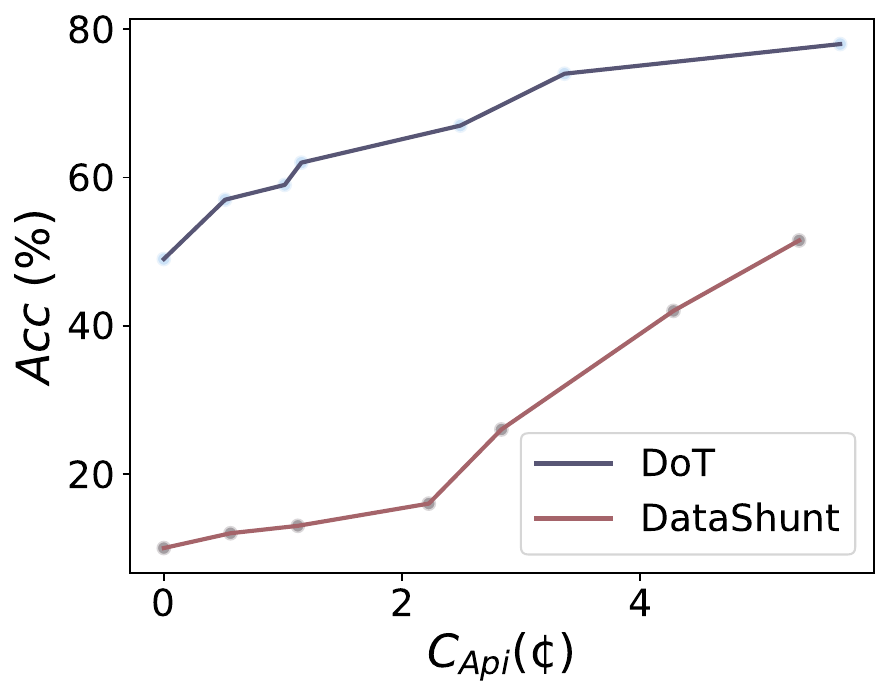}}
        \hfill
        \subfigure[CHAMP]{\includegraphics[width=0.49\linewidth]{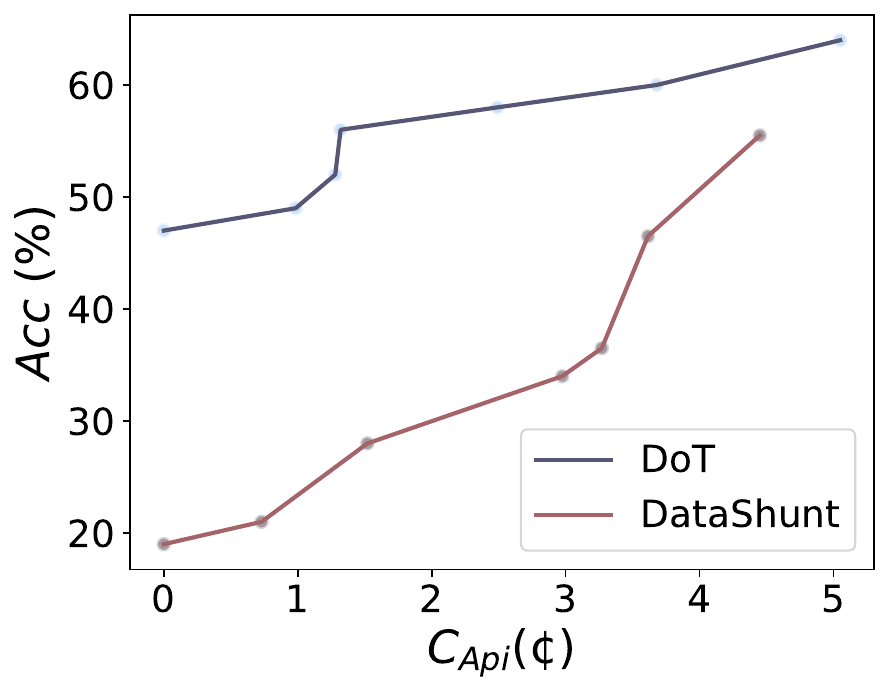}}
        \vskip 0.1em 
        \subfigure[P3]{\includegraphics[width=0.49\linewidth]{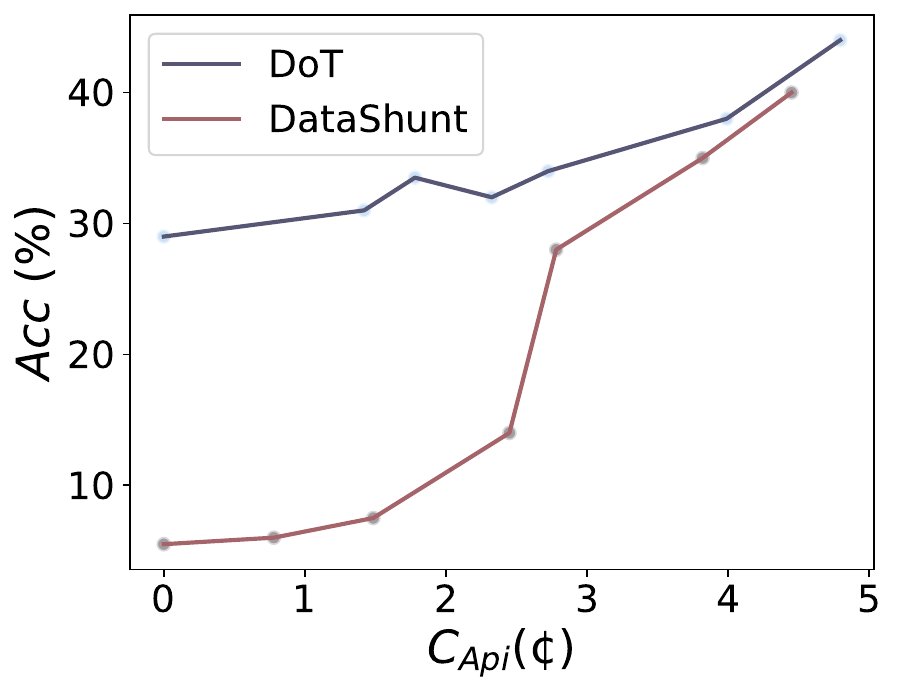}}
        \hfill
        \subfigure[CSQA]{\includegraphics[width=0.49\linewidth]{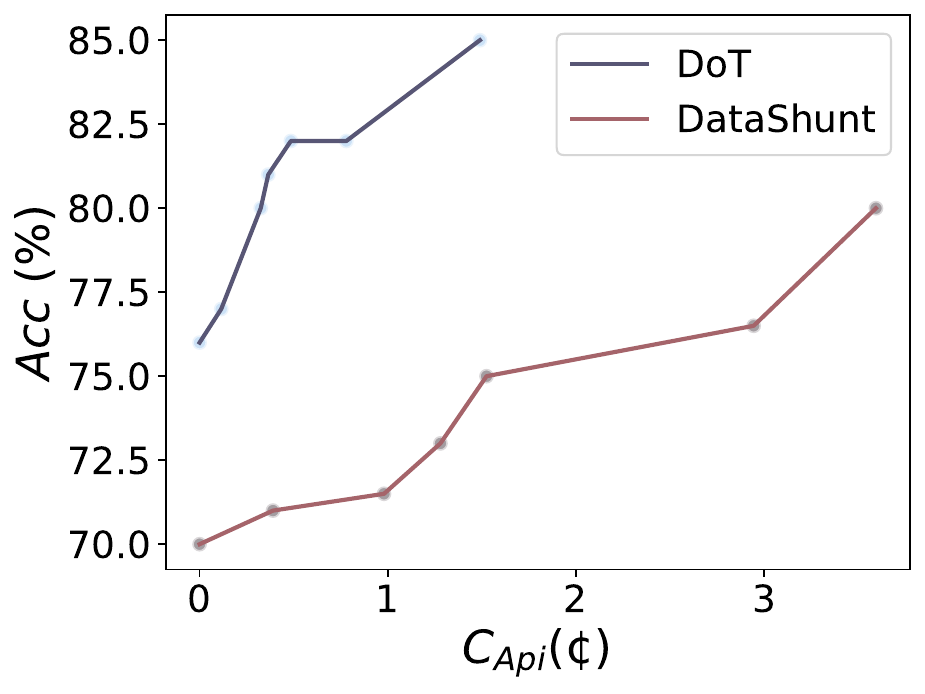}}
        \\
        \textbf{$\textbf{Acc}-\textbf{C}_{\textbf{API}}$ trade-off}
    }
    \hspace{2em} 
    \parbox[b]{0.47\linewidth}{
        \centering
        \subfigure[MATH]{\includegraphics[width=0.49\linewidth]{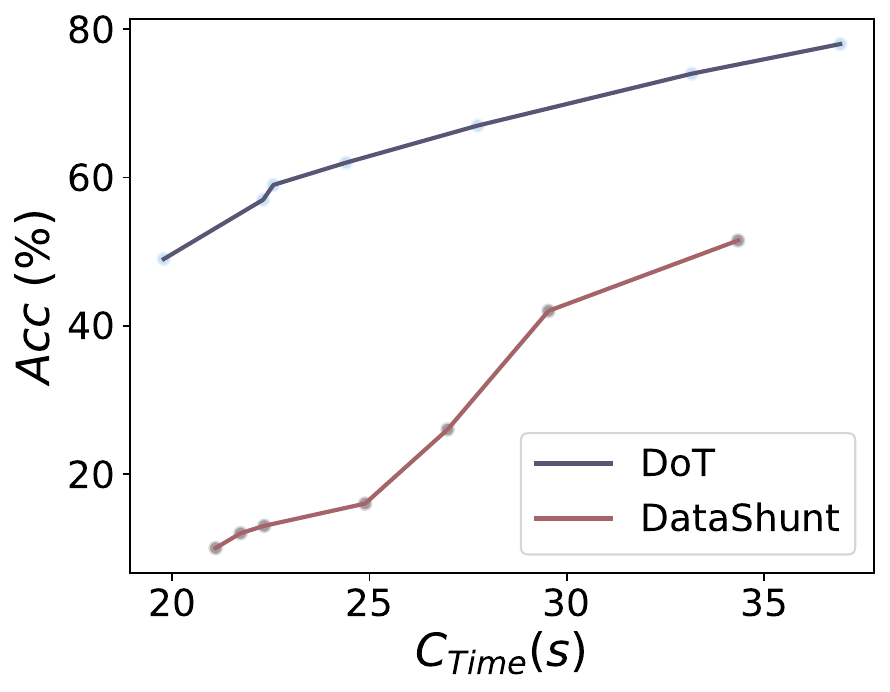}}
        \hfill
        \subfigure[CHAMP]{\includegraphics[width=0.49\linewidth]{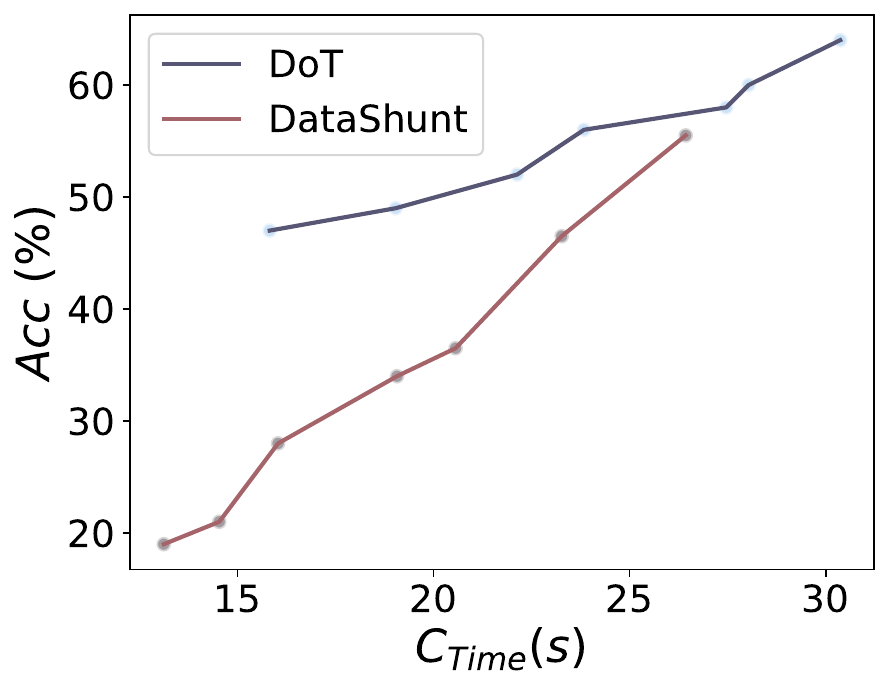}}
        \vskip 0.1em 
        \subfigure[P3]{\includegraphics[width=0.49\linewidth]{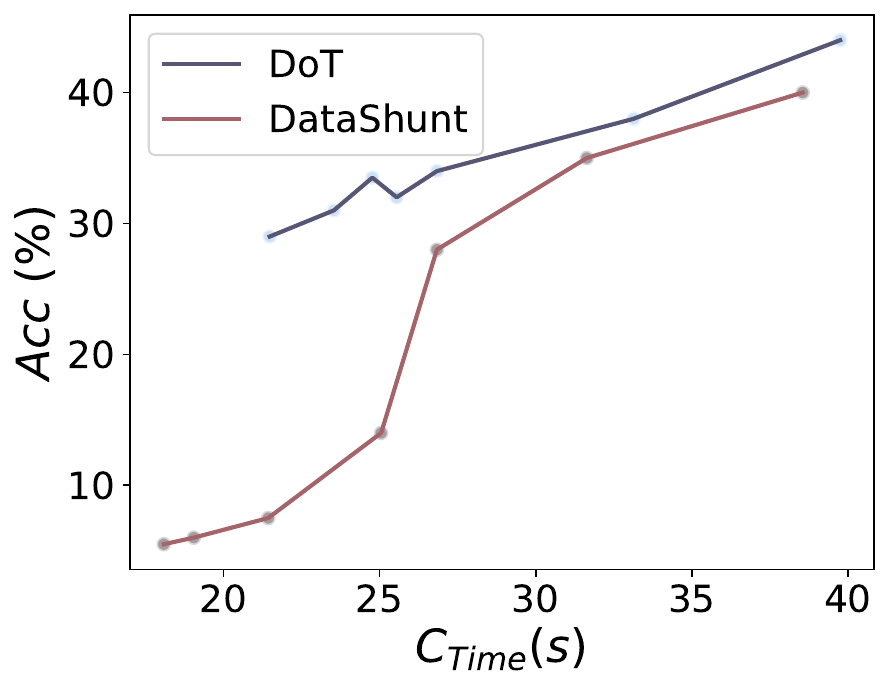}}
        \hfill
        \subfigure[CSQA]{\includegraphics[width=0.49\linewidth]{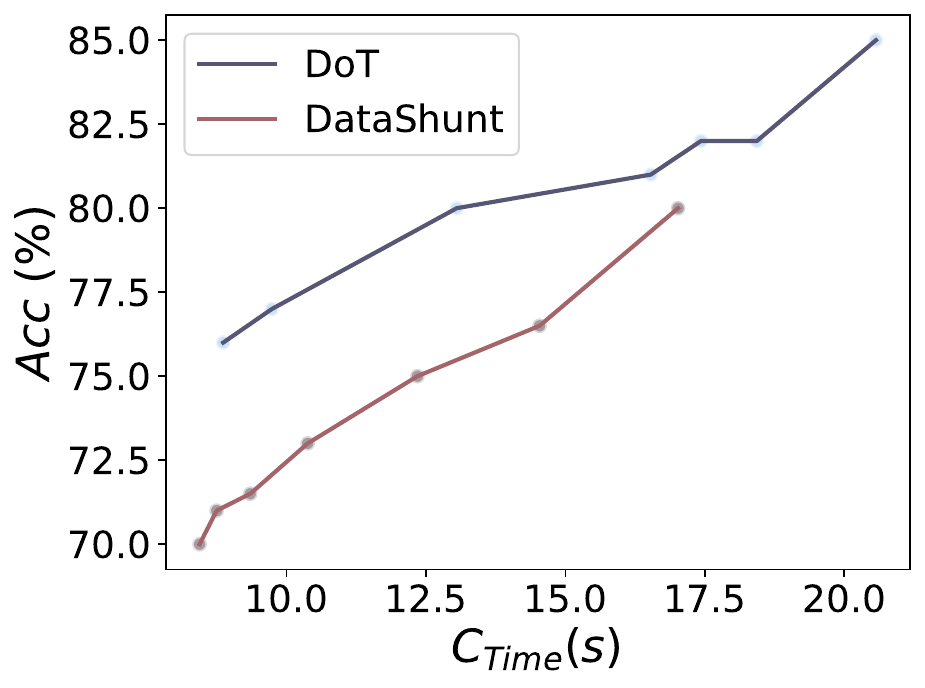}}
        \\
        \textbf{$\textbf{Acc}-\textbf{C}_{\textbf{Time}}$ trade-off}
    }

    \vspace{-3mm}
    \caption{$\textbf{Acc}-\textbf{Cost}$ trade-off curves on 4 benchmarks.}
    \label{exp: trade-off}
    \vspace{-4mm}

\end{figure*}

\vspace{-2mm}
\subsection{Efficiency \& Quality of Dataset Construction}

To validate the effectiveness of our method for difficulty assessment and optimal allocation search in dataset construction, we compared our $\alpha$-tree with other baseline search methods. We establish two baseline methods: \textbf{Zero-shot LLM}: This approach utilizes a LLM to assess the difficulty of sub-tasks. The evaluation model employs GPT-4o, utilizing a question-and-answer format that incorporates the task content, all sub-tasks, and the current sub-task into the prompt. \textbf{Binary-Search}: This method employs a binary search strategy. Initially, all sub-tasks are processed uniformly using GPT-4o. If the reasoning results are correct, half of the sub-tasks are randomly assigned to Llama 3-8B, with this random assignment repeated N times until successful. In the second round of binary search, half of the remaining sub-tasks assigned to the larger model are again randomly allocated to the smaller model. This process continues until all sub-tasks are either handled by the smaller model or incorrect reasoning persists. We set N to 5.

For our $\alpha$-tree approach, which sequentially searches for optimal solutions according to difficulty, we tested with n=1 and n=2 (where n represents the number of models whose allocation is altered at each search step). The comparison metrics included two quality assessment indicators: \textit{SLM Ratio}: the proportion of the smaller on-device model in the final allocation scheme, and \textit{SR}: success rate of reasoning with the final allocation scheme applied. Additionally, we considered two cost metrics: \textit{\# Evaluation}: each evaluation requires a complete reasoning pass over all sub-tasks based on the allocation strategy, and \textit{API Cost}, which refers to the expense incurred when invoking the cloud-based LLM API during dataset construction.

We visualize the quality and cost of the constructed dataset on the MATH benchmark. As shown in Table \ref{tbl:allocation}, in terms of construction quality, our $\alpha$-tree achieved the highest proportion of small model usage while maintaining the highest reasoning accuracy. This demonstrates that our method finds a more efficient collaborative allocation strategy, ensuring high-quality reasoning while minimizing inference costs. \textit{Zero-shot LLM} resulted in the lowest proportion of SLM usage, indicating that LLMs fail to assess the difficulty of sub-tasks. \textit{Binary-Search} also showed a lower SLM proportion compared to ours. While increasing the number of search attempts (N) could potentially yield better results through random search, this unordered and purely random strategy will incur significant costs. Even with the current 5-time search, the cost is already several times higher than our approach. Further scaling will bring unsustainable cost. Naturally, \textit{LLM-Eval} has the lowest cost, but its construction quality is significantly inferior.

\vspace{-4mm}

\subsection{Ablation Study}
\label{exp:ablation}

In this ablation study, we demonstrate the impact of the graph structure and task allocation mechanisms on the efficiency and effectiveness of our DoT model. Upon removing the graph structure, a sequential reasoning framework is used instead. In the absence of task allocation, a single language model or simple referral strategy is adopted. As shown in Table \ref{tbl:ablation}, the full DoT model, which includes both components, achieves high accuracy across various tasks while maintaining low time and API costs. When the graph component is removed, both accuracy and efficiency are reduced, indicating the impact of graph structure on enhancing reasoning performance and the effectiveness of graph-based parallel execution.

Additionally, removing the task allocation mechanism leads to higher API costs and increased time without substantial accuracy gains. For example, in the P3 task, removing allocation and solely using GPT-4o incurs a cost of 4.80\textcent compared to DoT’s 1.58\textcent, despite only a modest increase in accuracy (43\% vs. 41\%). This demonstrates that the allocation mechanism helps balance accuracy with efficiency.


Moreover, We are surprised to discover that, without considering costs, using the cloud-based LLM exclusively can greatly improve the reasoning accuracy on the challenging MATH and CHAMP benchmarks, where accuracy increased from 63\% to 78\% and from 57\% to 64\%, respectively, compared to the best baseline.

\vspace{-3mm}
\subsection{Trade-off Between Accuracy and Cost}

The ablation study demonstrates that our on-graph reasoning method significantly enhances the reasoning capability of LLMs, though at a higher computational cost. In this section, we delve into the detailed relationship between cost and reasoning accuracy to identify the achievable performance upper bound under different budget constraints. 
Based on our original task allocation strategy, we generate diverse model allocation schemes by proportionally assigning more sub-tasks to either the cloud-based LLM or the local deployed SLM according to the difficulty level of sub-tasks, resulting in the various data points shown in the figure. 
We also introduce \textit{DataShunt} as a baseline, where the proportion of tasks handled by the LLMs is adjusted by tuning the LLM evaluation thresholds. 
We test our method on four benchmarks, plotting the trade-off curves for accuracy versus API cost and accuracy versus time cost. 
As illustrated in Figure \ref{exp: trade-off}, our curve consistently remains above the baseline curve. Under the same accuracy, both the token cost and time expenditure of DoT are significantly lower than those of DataShunt, demonstrating that the model allocation method of DoT is quite efficient. Furthermore, as the budget increases, our DoT approach achieves a higher performance upper bound.













\vspace{-2mm}

\section{Conclusion}

In conclusion, our proposed \textit{Division-of-Thoughts} collaborative reasoning framework effectively enhances the reasoning performance of LLMs on edge devices by intelligently distributing tasks between local and cloud resources. \textit{DoT} not only breaks down complex tasks into manageable subtasks but also leverages a directed-acyclic graph to optimize the reasoning path, facilitating parallel edge-cloud processing. Furthermore, the plug-and-play adapter accurately assesses the difficulty of sub-tasks for allocation, ensuring high reasoning performance while significantly reducing costs of money and time.
Looking ahead, we aim to apply our reasoning framework across a broader and more diverse range of scenarios, striving to contribute to the advancement of edge AI.


\begin{acks}
  This work is supported in part by the National Natural Science Foundation of China under 23IAA02114 and 62472241, in part by the joint project of Infinigence AI \& Tsinghua University.
\end{acks}

\clearpage



\vspace{-4mm}
\section*{Appendix}
 
\appendix

\section{Supplementary Experiments}

\subsection{Generalizability of Adapter}
We conduct two supplementary experiments to evaluate the adapter's generalization performance across different tasks. 
\textbf{Intra-category generalization}: Within three mathematical benchmarks, train on two and test on the remaining one. 
\textbf{Cross-category generalization}: Train on six benchmarks and test on the remaining one. 
We also evaluate the performance of the adapter in the absence of any training.
Results are shown in Table~\ref{appendix:generalizability}, demonstrating that intra-category generalization within math-related benchmarks is almost as effective as task-specific training. And although there is a slight performance decrease in cross-category application, it still outperforms no training.

\begin{table}
\centering
 
\subtable[Intra-category generalization: MATH]{
\begin{tabular}{c|cccc}
Train       & Test & Acc  & Time & API    \\ 
\hline
MATH        & MATH & 59\% & 22.6 & 1.02\textcent  \\
DROP, CHAMP & MATH & 58\% & 26.9 & 1.28\textcent  \\
/           & MATH & 32\% & 33.2 & 2.18\textcent 
\end{tabular}
\label{tbl:gene_1}
}
 
\qquad
 
\subtable[Intra-category generalization: CHAMP]{        
\begin{tabular}{c|cccc}
Train      & Test  & Acc    & Time & API    \\ 
\hline
CHAMP      & CHAMP & 58\%   & 16.1 & 0.84\textcent  \\
MATH, DROP & CHANP & 56.5\% & 20.2 & 1.32\textcent  \\
/          & CHANP & 27\%   & 28.4 & 2.46\textcent 
\end{tabular}
\label{tbl:gene_2}
}

\qquad

\subtable[Intra-category generalization: DROP]{        
\begin{tabular}{c|cccc}
Train       & Test & Acc    & Time & API    \\ 
\hline
DROP        & DROP & 85\%   & 4.9  & 0.32\textcent  \\
MATH, CHAMP & DROP & 81.0\% & 6.4  & 0.43\textcent  \\
/           & DROP & 72.0\% & 7.2  & 0.49\textcent 
\end{tabular}
\label{tbl:gene_3}
}

\qquad

\subtable[Cross-category generalization: CSQA]{        
\centering
\renewcommand\arraystretch{1}
\setlength{\tabcolsep}{0.5mm}
\resizebox{\linewidth}{!}
{
\begin{tabular}{c|cccc}
Train                           & Test & Acc  & Time & API    \\ 
\hline
CSQA                            & CSQA & 82\% & 9.9  & 0.49\textcent  \\
DROP,CHAMP,MATH.P3,SCAN,Webshop & CSQA & 76\% & 12.5 & 0.58\textcent  \\
/                               & CSQA & 70\% & 15.2 & 0.84\textcent 
\end{tabular}
}
\label{tbl:gene_4}
}

\qquad

\subtable[Cross-category generalization: MATH]{        
\centering
\renewcommand\arraystretch{1}
\setlength{\tabcolsep}{0.5mm}
\resizebox{\linewidth}{!}
{
\begin{tabular}{c|cccc}
Train                           & Test & Acc  & Time & API    \\ 
\hline
MATH                            & MATH & 59\% & 22.6 & 1.02\textcent  \\
DROP,CHAMP,P3,SCAN,Webshop,CSQA & MATH & 53\% & 29.1 & 1.34\textcent  \\
/                               & MATH & 32\% & 33.2 & 2.18\textcent 
\end{tabular}
}
\label{tbl:gene_5}
}

\qquad

\subtable[Cross-category generalization: CHAMP]{        
\centering
\renewcommand\arraystretch{1}
\setlength{\tabcolsep}{0.5mm}
\resizebox{\linewidth}{!}
{
\begin{tabular}{c|cccc}
Train                          & Test  & Acc  & Time & API    \\ 
\hline
CHAMP                          & CHAMP & 58\% & 16.1 & 0.84\textcent  \\
DROP,MATH,P3,SCAN,Webshop,CSQA & CHAMP & 50\% & 19.6 & 1.35\textcent  \\
/                              & CHAMP & 27\% & 28.4 & 2.46\textcent 
\end{tabular}
}
\label{tbl:gene_6}
}

\qquad

\subtable[Cross-category generalization: P3]{        
\centering
\renewcommand\arraystretch{1}
\setlength{\tabcolsep}{0.5mm}
\resizebox{\linewidth}{!}
{
\begin{tabular}{c|cccc}
Train                             & Test & Acc  & Time & API    \\ 
\hline
P3                                & P3   & 41\% & 23.5 & 1.58\textcent  \\
DROP,CHAMP,MATH,SCAN,Webshop,CSQA & P3   & 33\% & 29.3 & 1.73\textcent  \\
/                                 & P3   & 15\% & 35.1 & 2.38\textcent 
\end{tabular}
}
\label{tbl:gene_7}
}

\caption{Evaluation of Generalizability}
\label{appendix:generalizability}
\vspace{-10mm}
\end{table}

\vspace{-2mm}
\subsection{Evaluation of Task Decomposition Quality}

For the quality of the task decomposition, we compare our method with a vanilla prompting strategy, focusing on evaluating the \textit{independence} of subtasks and the impact of the decomposition strategy on both the final reasoning accuracy and the reasoning cost. 
For \textit{independence}, we aim to determine whether the generated subtasks are independent (i.e., without overlapping mathematical computation) and calculate the independence rate for performance ealuation. We first manually annotate the independence of the decomposition results for 50 tasks in the MATH benchmark. Our method achieves 91.3\% in independence rate, significantly outperforming the vanilla prompting strategy, which only reaches 74.2\%. This demonstrates the high quality of our task decomposition approach. 

\begin{table}

\centering
 
\subtable[Decomposition Independence Evaluation on CSQA]{
\centering
\label{tbl:indep_csqa}
\renewcommand\arraystretch{1}
\setlength{\tabcolsep}{0.5mm}
\resizebox{\linewidth}{!}
{
\begin{tabular}{c|c|c|c|c}
Method                          & Independence & Acc    & Time & API    \\ 
\hline
(Ours) DoT decomposition        & 95\%         & 82\%   & 9.9 & 0.49\textcent \\
vanilla prompting decomposition & 87\%         & 81\% & 13.2 & 0.77\textcent 
\end{tabular}
}
}
 
\qquad
 
\subtable[Decomposition Independence Evaluation on MATH]{        
\centering
\label{tbl:indep_math}
\renewcommand\arraystretch{1}
\setlength{\tabcolsep}{0.5mm}
\resizebox{\linewidth}{!}
{
\begin{tabular}{c|c|c|c|c}
Method                          & Independence & Acc    & Time & API    \\ 
\hline
(Ours) DoT decomposition        & 90\%         & 59\%   & 22.6 & 1.02\textcent \\
vanilla prompting decomposition & 76\%         & 57\% & 38.4 & 1.78\textcent 
\end{tabular}
}
}

\qquad

\subtable[Decomposition Independence Evaluation on P3]{        
\centering
\label{tbl:indep_p3}
\renewcommand\arraystretch{1}
\setlength{\tabcolsep}{0.5mm}
\resizebox{\linewidth}{!}
{

\begin{tabular}{c|c|c|c|c}
Method                          & Independence & Acc    & Time & API    \\ 
\hline
(Ours) DoT decomposition        & 86\%         & 41\%   & 23.5 & 1.58\textcent  \\
vanilla prompting decomposition & 71\%         & 36.5\% & 24.5 & 2.23\textcent 
\end{tabular}
}
}

\caption{Evaluation of Decomposition Independence}
\label{tbl:indepedence}
\vspace{-6mm}
\end{table}

Subsequently, we employ GPT-4o to evaluate the independence of subtasks generated by both methods and compared its results to manual annotations. The GPT-4o assessments achieve a 93\% consistency with human annotations, indicating that GPT-4o can provide annotation quality comparable to that of humans and can be effectively used for large-scale evaluations. 
Therefore, we utilize GPT-4o to evaluate the independence of task decomposition across the entire test set, with the results presented in Table~\ref{tbl:indepedence}. The results indicate that our task decomposition method achieves an average improvement of 15.7\% in subtask independence compared to the baseline, demonstrating a clear advantage. Moreover, this improvement contributes to enhanced reasoning accuracy and reduced reasoning costs.



\vspace{-2mm}
\subsection{Impact of the Number of Sub-tasks on Reasoning Efficiency}
\begin{table}
\centering
\renewcommand\arraystretch{1}
\setlength{\tabcolsep}{0.5mm}
\resizebox{\linewidth}{!}
{

\begin{tabular}{c|c|c|c|c}
\begin{tabular}[c]{@{}c@{}}\textbf{}\\\end{tabular} & \multicolumn{2}{c|}{P3}                   & \multicolumn{2}{c}{MATH}                   \\ 
\hline
\textbf{\# Subtask}                                 & \textbf{↑Accuracy} & \textbf{↓Cost\_time} & \textbf{↑Accuracy} & \textbf{↓Cost\_time}  \\
3                                                   & 14.5\%             & 18.3\%               & 9\%                & 13.5\%                \\
4                                                   & 18\%               & 21.7\%               & 13\%               & 15.5\%                \\
5                                                   & 20\%               & 25.4\%               & 14\%               & 17.2\%                \\
6                                                   & 13\%               & 22.6\%               & 17\%               & 18.4\%                \\
7                                                   & 14\%               & 23.9\%               & 15\%               & 17.6\%                \\
8                                                   & 12\%               & 28.7\%               & 13\%               & 25.7\%                \\
>8                                                   & 17.5\%             & 32.1\%               & 18\%               & 30.3\%              

\end{tabular}
}
\caption{Evaluation of DoT's Scalibility}
\label{tbl:scale}
\vspace{-10mm}
\end{table}

As problems become more complex and the number of decomposed sub-tasks increases, whether our framework can achieve a significant improvement in reasoning efficiency while maintaining accuracy remains a valuable research question.
Therefore, we conduct supplementary experiment, where we compare our method with baseline that \textit{sequentially reason all sub-tasks on SLM} and perform statistical analyses based on varying numbers of sub-tasks. The results in Table~\ref{tbl:scale} demonstrate that our model consistently maintains high reasoning efficiency across tasks with different numbers of sub-task nodes.

Furthermore, as the number of sub-tasks increases, the enhancement in efficiency becomes increasingly pronounced. When the number of subtasks exceeds eight, the reduction in reasoning time surpasses even 30\%. This outcome is inherently logical, for as the quantity of sub-tasks escalates, the task graph grows in complexity, thereby expanding the potential for parallel reasoning and amplifying the advantages in terms of reasoning time.

\section{Benchmarks and Implementation Details}

\subsection{Mathematics}

\begin{itemize}
    \item \textbf{MATH}~\cite{hendrycks2021measuring}. Mathematics Aptitude Test of Heuristics (MATH), comprises 12,500 problems from prestigious U.S. mathematics competitions like AMC and AIME. These problems, collected from platforms such as AoPS, test advanced problem-solving skills beyond standard K-12 math. Each problem includes a step-by-step solution and a final answer. The dataset spans seven subjects: Prealgebra, Algebra, Number Theory, Counting and Probability, Geometry, Intermediate Algebra, and Precalculus, with difficulty levels ranging from 1 (easy) to 5 (challenging), allowing models to learn and apply various mathematical heuristics.
    
    \item \textbf{CHAMP}~\cite{mao2024champ}. Concept and Hint-Annotated Math Problems (CHAMP). This benchmark features 270 non-routine, competition-level problems sourced from Engel's Problem-Solving Strategies. The problems span five categories: number theory, polynomial, sequence, inequality, and combinatorics, requiring creative strategies and specific tricks. Each problem includes a final checkable answer and a step-by-step solution in natural language. The dataset contains 54 concepts and 330 hints, averaging 1.4 concepts, 1.7 hints, and 6 solution steps per problem. Problem statements average 20.2 words, with solutions averaging 10.9 words per step, highlighting the dataset's complexity and challenge.
    
    \item \textbf{DROP}~\cite{dua2019drop}. Discrete Reasoning Over Paragraphs (DROP), is an English reading comprehension dataset with 96k adversarially-created questions. It challenges systems to resolve references in a question and perform discrete reasoning operations like addition, counting, or sorting over multiple input positions within a paragraph. The dataset is crowdsourced and derived from Wikipedia passages designed for complex questions. DROP demands a deeper understanding of paragraph content than previous datasets, with rigorous validation to ensure the quality of its development and test sets.
\end{itemize}

\subsection{Logic}
\begin{itemize}
    \item \textbf{P3}~\cite{schuster2021programming}. Python Programming Puzzles (P3), introduces a new type of programming challenge for evaluating program synthesis. P3 contains 397 Python-based puzzles, where the goal is to find an input that makes a given function return True. The puzzles span various difficulty levels, from simple string manipulations to complex algorithmic problems. 
    \item \textbf{Scan}~\cite{lake2018generalization}. This benchmark is used to evaluate the sequence-to-sequence learning ability to translate simplified natural language commands into action sequences. SCAN contains 20,910 commands generated by a phrase-structure grammar, which describe basic actions such as "jump" or "walk" and their combinatorial variations (e.g., "jump around left"). The logical structure and constraints involved in the translation process make SCAN ideal for assessing LLM reasoning capabilities.
\end{itemize}

\vspace{-2mm}
\subsection{Commonsense}
\begin{itemize}
    \item \textbf{COMMONSENSEQA}~\cite{talmor2018commonsenseqa}. This benchmark is a challenging dataset designed to test commonsense question answering. It consists of 12,247 multiple-choice questions created using concepts from CONCEPTNET. Each question is authored by crowd-workers to differentiate between multiple target concepts that share a semantic relation with a source concept, encouraging the use of prior knowledge and complex reasoning. 
\end{itemize}

\vspace{-2mm}
\subsection{On-device AI Assistant Application}
\begin{itemize}
    \item \textbf{Webshop}~\cite{yao2022webshop}. WebShop serves as a simulated e-commerce environment featuring 1.18 million real-world products and 12,087 crowd-sourced text instructions. Designed to evaluate language agents, it challenges them to navigate diverse web pages and perform actions based on natural language product specifications. Agents encounter obstacles such as interpreting compositional instructions, reformulating queries, and understanding noisy text on webpages, while strategically exploring to fulfill product requirements. The modular design of WebShop separates website navigation from task-specific elements, allowing for easy adaptation to new tasks and domains. This dataset provides a robust platform for assessing the capabilities of language agents in an interactive, real-world-inspired setting, emphasizing their ability to comprehend and act on complex instructions.
\end{itemize}

\vspace{-2mm}
\subsection{Implementation Details across Benchmarks}
\label{appendix:imple_details}
\textbf{MATH.}
For MATH tasks, we employ a structured workflow comprising four key stages: Task Decomposition, Model Allocation, Dependency Graph Construction, and Step-by-Step Reasoning Based on the Graph.
In the task decomposition phase, we prompt the LLM with exemplars of manually decomposed complex problems. The LLM is then instructed to generate manageable subtasks that collectively solve the primary challenge. Subsequently, we task the LLM with establishing subtask dependencies, where a relationship $Step_i \rightarrow Step_j$ indicates that $Step_i$ must precede $Step_j$.
Using the derived dependencies, we construct a reasoning graph via Breadth-First Search (BFS). Subtasks at the same depth are processed in parallel. Upon completion of all subtasks, we conduct a final query to obtain the ultimate solution. This solution undergoes LLM-based evaluation through comparison with the ground truth.

\textbf{CHAMP.}
For CHAMP tasks, the process is largely similar to the MATH process with the primary distinction being the specific few-shot examples of human-written decompositions provided.

\textbf{DROP.}
For DROP tasks, we adapt our approach to accommodate the format of questions based on given texts. The prompt for each step incorporates relevant background information provided in the dataset while maintaining the core implementation structure used in other benchmarks.

\textbf{P3.}
The Programming Puzzle tasks present a unique challenge, requiring the LLM to generate inputs that yield a 'True' output for a given function. In addition to the puzzle description, we provide the LLM with the expected data type of the final input. The evaluation process for P3 differs from other benchmarks: we execute the program using the LLM-generated input and assess the correctness based on the program's output.

\textbf{SCAN.}
For SCAN tasks focused on translating natural language instructions into action sequences, the steps until evaluation stay the same. The final phase involves converting the model's natural language outputs into standardized action sequences using few-shot examples. The evaluation is conducted by directly comparing these sequences with the true answers, ensuring the outputs accurately match the expected actions.

\textbf{COMMONSENSEQA.}
For CSQA tasks, presented as multiple-choice questions based on common sense, our implementation has an identical structure as other tasks, while both the problem and its options are presented to the LLM for the following task decomposition and reasoning. The Final Evaluation consists of the LLM choosing the most plausible answer from the provided options. This selected answer is cleaned and will directly compare with the correct choice.

\textbf{WebShop.} 
The WebShop task presents unique challenges due to its interactive nature, where each action influences subsequent states and available options. Unlike our other experiments, WebShop is not perfect for the construction of a complete, predefined reasoning graph. Instead, our framework dynamically generates and executes sub-tasks based on the current state of the shopping session, in which the shopping process is divided into high-level sub-tasks/thoughts. We utilized the Model Allocation for each step when generating an action, either be `think[]`, `click[]`, or `plan[]`. The first step would be to generate a keyword from the instructions to search on the website. Then, the top \textit{N = 10} matched items are recorded for the following procedures. With the conversation from the webpage, we implemented the task decomposition process to get a roadmap for the following actions. Prompting each step at once to the LLMs, the detailed information about each item will be recorded in the search history. Then, an evaluation and comparison process is prompted to choose the best item from the list. Lastly, the LLM will be prompted dynamically based on the previous steps, and self-navigated to complete the final purchase. This implementation demonstrates our framework's flexibility in handling tasks with dynamic, state-dependent decision-making processes.

\vspace{-3mm}
\section{Prompts}
\subsection{Task Decomposition}
I will now give you a [Based on the type of problem]. The type of problem is {type}. Please break this problem down into several easy-to-solve steps.

1 examples are as follows:
[Manual Written Examples]

Now the command is {question}, please decompose it into easy-to-solve steps like the examples.
Answer Format: (Please write each broken-down question step on a separate line, starting with a number.)

To solve the question "xxx", we need to know:
"1. question $step_1$",
"2. question $step_2$",
"3. question $step_3$".
...

\vspace{-2mm}
\subsection{Dependency Construction}
\textit{System Prompt:}\\
Now we have a problem, which we have broken down into many sub-problems. I want you to understand the connection between these sub-problems\\
\textit{User Prompt:\\}
    The init problem is {question}. And the sub-problems are {steps}. Please provide your understanding of the relationships between these sub-problems. Your response must be concise.

    Now we need to create standardized connections for the relationships between these sub-problems.
    Now Given the following subtasks for question: [question], determine the dependencies between them:

    [List of Steps]
        
    Please list the dependencies in the format 'Subproblem A [xxx] -> Subproblem B [xxx]' indicating that Sub-problem A must be completed before Sub-problem B can start.
    Please identify any potential conditional dependencies from a logical perspective.
    
    Answer format: (Please strictly follow the format. Each dependency should be separated by a new line. No explanation is required.)\\
    Step $ID_i$ [ sub-problem i ] -> Step $ID_j$ [ sub-problem j ]\\
    Step $ID_j$ [ sub-problem m ] -> Step $ID_n$ [ sub-problem n ] ...
\vspace{-2mm}
\subsection{Subtask Reasoning}
\textit{System Prompt:}\\
Here is a math word problem. I will first provide a passage of the problem to set the context. Then, I will ask a specific question that requires you to use the information from the problem description, along with calculation and reasoning, to solve it. \\
Passage:
[passage] Question: [question]

I have broken this math question down into several smaller questions. I will assign you sub-questions one by one, and provide the results of previous sub-questions as a reference for your reasoning.
Please solve the question according to mathematical logic.

\textit{For each steps}
So far, the answers to the resolved sub-questions are as follows: The format is Sub-question-Id: xxx; Sub-question: xxx; Answer: xxx.
Sub-question-Id: [Corresponding ID]; Sub-question: [Corresponding Step]; Answer: [Corresponding Solution for the step]\\
Among them, sub-questions {predecessors} are directly related to this sub-question, so please pay special attention to them.
The sub-question to solve now is xxx: \{subtask\}
Based on the information above, please provide a concise and clear answer

\end{document}